%% file: main.tex
\documentclass[10pt,twocolumn,letterpaper]{article}
\usepackage[utf8]{inputenc}
\usepackage[T1]{fontenc}
\usepackage{multirow}
\usepackage{iccv}
\usepackage[dvipsnames]{xcolor}
\usepackage[linesnumbered,ruled,vlined]{algorithm2e}
\usepackage{times}
\usepackage{epsfig}
\usepackage{graphicx}
\usepackage{amsmath}
\usepackage{amssymb}
\usepackage[toc,page]{appendix}
\usepackage{graphicx}
\usepackage{caption}
\usepackage{float}

\usepackage{wrapfig}
\usepackage{subcaption}
\usepackage[export]{adjustbox}
\usepackage{booktabs}
\usepackage{url}
\usepackage[autostyle]{csquotes}
\usepackage{epigraph}
\usepackage{graphicx}
\usepackage{pifont}
\usepackage{authblk}
\usepackage[bb=boondox]{mathalfa}
\newcommand*\samethanks[1][\value{footnote}]{\footnotemark[#1]}
\newcommand\scalemath[2]{\scalebox{#1}{\mbox{\ensuremath{\displaystyle #2}}}}

\newcommand\blfootnote[1]{%
  \begingroup
  \renewcommand\thefootnote{}\footnote{#1}%
  \addtocounter{footnote}{-1}%
  \endgroup
}
\SetKwInput{KwInput}{Input}                
\SetKwInput{KwData}{Data}                
\SetKwInput{KwParams}{Parameters}                
\SetKwInput{KwOutput}{Output}              
\SetKwBlock{KwRepeat}{Repeat}              

\usepackage[breaklinks=true,bookmarks=false]{hyperref}

\iccvfinalcopy 
\def\iccvPaperID{3306} 

\ificcvfinal\fi

\begin{document}

\title{MixMo: Mixing Multiple Inputs for Multiple Outputs via Deep Subnetworks}
\author[1]{Alexandre Ramé\thanks{Equal contribution.}\thanks{Correspondence to alexandre.rame@lip6.fr}}
\author[1,2]{Rémy Sun\samethanks[1]}
\author[1,3]{Matthieu Cord}
\affil[1]{Sorbonne Université, CNRS, LIP6, Paris, France}
\affil[2]{Optronics \& Missile Electronics, Land \& Air Systems, Thales}
\affil[3]{Valeo.ai}
\renewcommand\Authands{ and }
\newcommand{\cmark}{\ding{51}}%
\newcommand{\xmark}{\ding{55}}%
\newcommand{\fix}{\marginpar{FIX}}
\newcommand{\new}{\marginpar{NEW}}
\newcommand\mycommfont[1]{\footnotesize\ttfamily\textcolor{blue}{#1}}
\maketitle
\ificcvfinal\thispagestyle{empty}\fi
\begin{abstract}%
\input{sections/00_abstract.tex}
\end{abstract}%
\blfootnote{Proceedings of IEEE International Conference on Computer Vision (ICCV), 2021}
\input{sections/01_introduction_arxiv.tex}
\input{sections/02_relatedwork_arxiv.tex}

\input{sections/03_model.tex}
\input{sections/04_experiments.tex}
\input{sections/05_conclusion.tex}

\subsubsection*{Acknowledgments}
This work was performed using HPC resources from GENCI-IDRIS (Grant 2021-AD011012262), with financial supports from the ANR agency in the chair VISA-DEEP (ANR-20-CHIA-0022-01),
and from Rémy's CIFRE grant between Thales Land and Air Systems and Sorbonne University.
We thank Andrei Bursuc for his detailed feedbacks.
{\small
\bibliographystyle{ieee_fullname}
\bibliography{main}
}

\clearpage

\section{Appendix}

\input{sections/06_appendix.tex}

\end{document}

%% file: sections/00_abstract.tex
Recent strategies achieved ensembling ``for free'' by fitting concurrently diverse
subnetworks inside a single base network. The main idea during training is that
each subnetwork learns to classify only one of the multiple inputs
simultaneously provided.
However, the question of how to best mix these multiple inputs has not been studied so far.

In this paper, we introduce MixMo, a new generalized framework for learning
multi-input multi-output deep subnetworks. Our key motivation is to replace the
suboptimal summing operation hidden in previous approaches by a more appropriate
mixing mechanism. For that purpose, we draw inspiration from successful mixed
sample data augmentations.
We show that binary mixing in features - particularly with rectangular patches from CutMix - enhances results by making subnetworks stronger and more diverse.

We improve state of the art for image classification on CIFAR-100 and Tiny ImageNet datasets.
Our easy to implement models notably outperform data augmented deep ensembles, without the inference and memory overheads.
As we operate in features and simply better leverage the expressiveness of large networks, we open a new line of research complementary to previous works.%

%% file: sections/01_introduction_arxiv.tex
\section{Introduction}%

Convolutional Neural Networks (CNNs) have shown exceptional performance in
computer vision tasks, notably classification \cite{krizhevsky2009learning}.
However, among other limitations, obtaining reliable predictions remains
challenging \cite{hendrycks2018benchmarking,ovadia2019can}. For additional
robustness in real-world scenarios or to win Kaggle competitions, CNNs usually
pair up with two practical strategies: data augmentation and ensembling.

Data augmentation reduces overfitting and improves generalization, notably by
diversifying training samples \cite{gontijo2020affinity}. Traditional approaches are label-preserving. In contrast, recent \textbf{mixed sample data
augmentation} (MSDA) create artificial samples by mixing multiple inputs and their labels
proportionally to a ratio $\lambda$. The seminal
work Mixup \cite{zhang2018mixup} linearly interpolates pixels while Manifold
Mixup \cite{manifoldmixup19} interpolates latent features in the network.
Binary masking MSDAs \cite{french2020milking,harris2020mix,kim2020puzzle} such as
CutMix \cite{yun2019cutmix} have since diversified mixed samples by pasting patches from one image onto another in place of interpolation.

\input{pictures/model/fig_intro_motivation.tex}%

Aggregating predictions from a diverse set of neural networks (\textit{i.e.} with different failure cases) strongly improves
generalization
\cite{dietterich2000ensemble,hansen1990neural,lakshminarayanan2016simple},
notably uncertainty estimation
\cite{ashukha2020pitfalls,gustafsson2020evaluating,ovadia2019can}.
An ensemble of several small networks usually performs
better than one large network empirically \cite{chirkova2020deep,lobacheva2020power}.
Yet, unfortunately, ensembling is costly in time and memory both at training and inference: this often limits applicability.%

In this paper, we propose \textbf{MixMo}, a new generalized multi-input multi-output
framework: we train a base network with $M\geq 2$ inputs and outputs. This way, we fit $M$ independent subnetworks
\cite{gao2019ntraensemble,havasi2020raining,soflaei2020aggregated} defined by an input/output pair and a subset of network weights. This is
possible as large networks only leverage a subset of their weights \cite{lottery2019}. Rather than pruning (ie, eliminating) inactive filters
\cite{e9ee2143e19d49cf9cbe8861950b6b2a,li2017pruning}, we seek
to fully use the available neurons and over parameterization through multiple subnetworks.


The challenge is to prevent homogenization and enforce diversity among
subnetworks with no structural differences. Thus, we consider $M$ (input,
label) pairs at the same time in training: $\{(x_i, y_i)\}_{0 \leq i < M}$. \textbf{$M$ images are treated simultaneously}, as shown on Fig.~\ref{fig:intro} with $M=2$. The $M$ inputs are encoded by $M$ separate convolutional layers $\{c_i\}_{0 \leq i < M}$ into a shared latent space before being mixed. The representation is
then fed to the core network, which finally branches out into $M$ dense layers $\{d_i\}_{0 \leq i < M}$.
Diverse subnetworks naturally emerge as $d_i$ learns to classify $y_i$ from
input $x_i$. At inference, the same image is repeated $M$ times: we obtain
ensembling ``for free'' by averaging $M$ predictions.%

The key divergent point between MixMo variants lies in the \textbf{multi-input
mixing block} that seeks features independence.
Should the merging be a basic summation or a concatenation, we would recover MIMO \cite{havasi2020raining} or respectively Aggregated Learning \cite{soflaei2020aggregated} - which both featured this multi-input multi-output strategy.

Our main intuition is simple: we see summing as a balanced and restrictive form
of Mixup \cite{zhang2018mixup} where $\lambda=\frac{1}{M}$. By analogy,
we draw from the considerable MSDA literature to design a more
appropriate mixing block. In particular, we leverage binary masking methods to
ensure subnetworks diversity. Our framework allows us to create a
new Cut-MixMo variant inspired by CutMix \cite{yun2019cutmix}, and illustrated in
Fig.~\ref{fig:intro}: a patch of features from the first input is pasted into
the features from the second input.

This asymmetrical mixing also raises new questions regarding information flow in
the network's features. We tackle the imbalance between the multiple
classification training tasks via a new weighting scheme.
Conversely, MixMo's double nature as a
new mixing augmentation in features yields important insights on traditional MSDA.

In summary, our contributions are threefold:%
\begin{enumerate}
\item We propose a general framework, MixMo, connecting two successful fields:
mixing samples data augmentations $\&$ multi-input multi-output ensembling.
\item We identify the appropriate mixing block to best tackle
  the diversity/individual accuracy trade-off in subnetworks: our easy to implement Cut-MixMo benefits from the synergy between CutMix and ensembling.%
\item We design a new weighting of the loss components to properly leverage the asymmetrical inputs mixing.
\end{enumerate}

\input{pictures/experiments/fig_main.tex}
We demonstrate excellent accuracy and uncertainty estimation with MixMo on CIFAR-10/100 and Tiny ImageNet.
Specifically, Cut-MixMo with $M=2$ reaches state of the art on these standard datasets: as exhibited by Fig.~\ref{fig:main}, it outperforms CutMix, MIMO and deep ensembles, at (almost) the same inference cost as a single network.%

%% file: pictures/model/fig_intro_motivation.tex
\begin{figure}%
\centering%
\includegraphics[width=\linewidth]{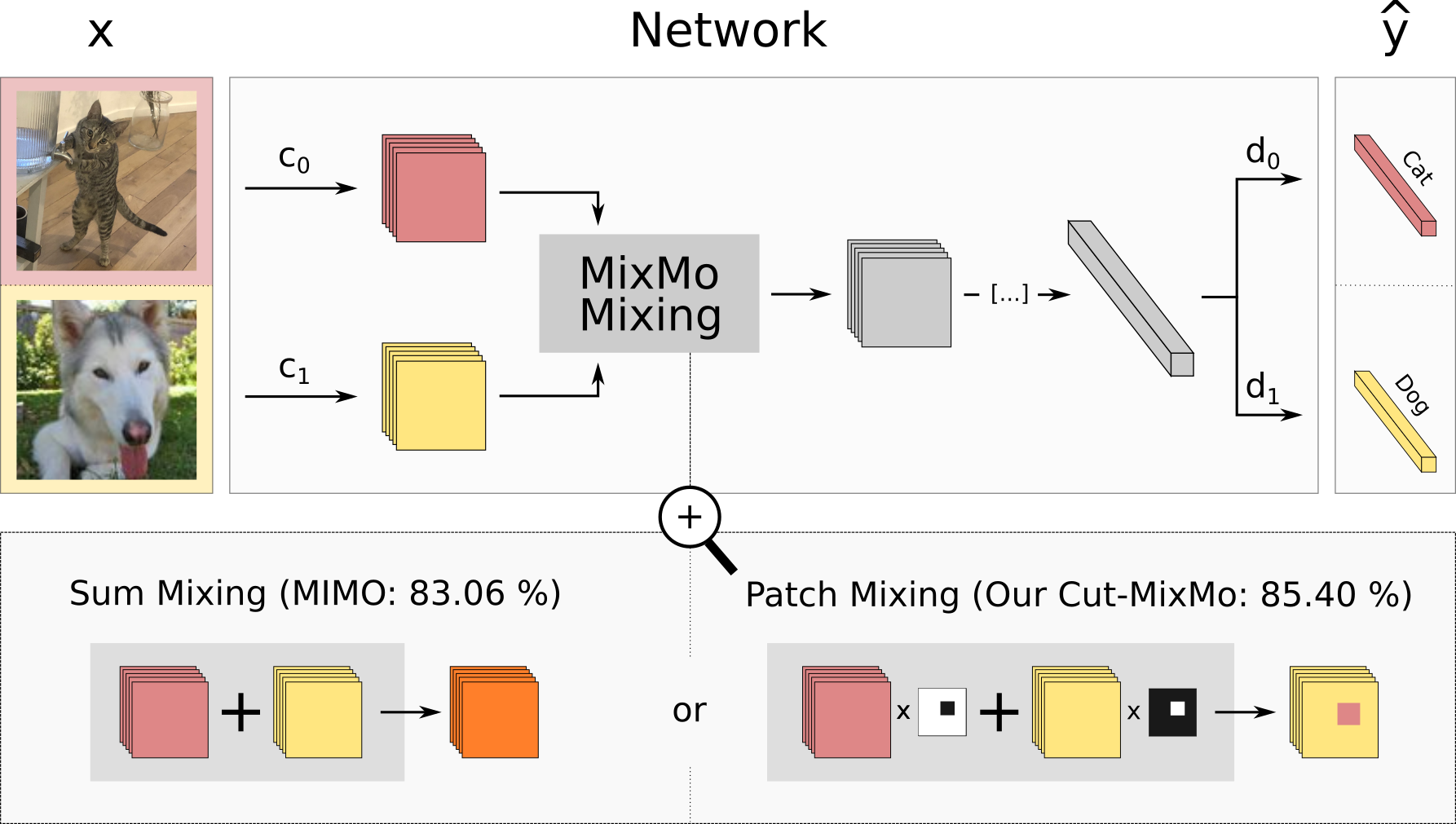}%
\caption{\textbf{MixMo overview}. We embed $M=2$ inputs into a shared space with convolutional layers ($c_1$, $c_2$), mix them, pass the embedding through further layers and output 2 predictions via dense layers ($d_1$, $d_2$). The key point of our MixMo is the mixing block. Mixing with patches performs better than basic summing: $85.40\%$ vs.\ $83.06\%$ (MIMO \cite{havasi2020raining}) on CIFAR-100 with WRN-28-10.}%
\label{fig:intro}%
\end{figure}%

%% file: pictures/experiments/fig_main.tex
\begin{figure}[!t]%
\centering%
\includegraphics[width=1.0\linewidth]{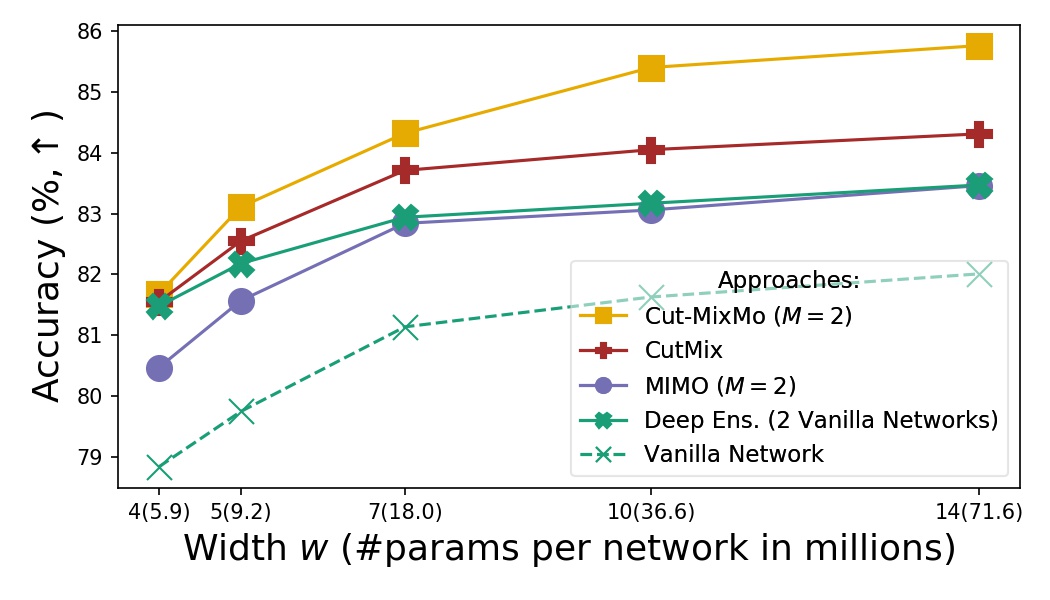}%
\caption{\textbf{Main results}. CIFAR-100 with WRN-28-$w$. Our Cut-MixMo variant (patch mixing and $M=2$) surpasses CutMix and deep ensembles (with half the parameters) by leveraging over-parameterization in wide networks.}%
\label{fig:main}%
\end{figure}%

%% file: sections/02_relatedwork_arxiv.tex
\section{Related work}

\subsection{Data augmentation}

\paragraph{}%
CNNs are known to memorize the training data \cite{45820} and make
overconfident predictions \cite{guo2017calibration} to the detriment of generalization on new test examples. \textbf{Data Augmentation} (DA)
inflates the training dataset's size by creating artificial samples from
available labeled data. Beyond slight perturbations (\textit{e.g.} rotation), recent works \cite{Cubuk_2020_CVPR_Workshops, hendrycks2019augmix} apply stronger transformations \cite{he19_data_augmen_revis}. CutOut \cite{devries2017improved} randomly deletes regions of
images in training and prevents models from focusing on a single pixels region,
similarly to how regularizations like Dropout
\cite{srivastava2014dropout} or DropBlock \cite{ghiasi2018dropblock} force
networks to leverage multiple features.%

\textbf{Mixed Sample Data Augmentation} (MSDA) recently expanded the notion of
DA. From pairs of labeled samples $\{(x_{i}, y_i), (x_{k}, y_{k})\}$, they
create virtual samples: $\left(m_x(x_{i}, x_{k}, \lambda), \lambda y_i +
  (1-\lambda)y_{k}\right)$ where $\lambda \sim \text{Beta}(\alpha,\alpha)$. 
\cite{liang2018understanding} shows that mixing the targets differently than this linear interpolation may cause
underfitting and unstable learning. Indeed, approaches mainly focus on developing the most effective input mixing $m_x$. In \cite{inoue18_data_augmen_by_pairin_sampl_images_class,Tokozume_2018_CVPR,tokozume2018learning,zhang2018mixup}, $m_x$ performs a simple linear interpolation between pixels: \textit{e.g} in Mixup \cite{zhang2018mixup}, $m_x(x_{i}, x_{k}, \lambda) = \lambda x_{i} + (1-\lambda) x_{k}$.  Theoretically, it regularizes outside the training distribution \cite{onmixup2020,guo2019mixup,zhang2020does} and applies label smoothing \cite{muller2019does,pereyra2017regularizing}.%

CutMix draws from Mixup and CutOut
\cite{devries2017improved} by pasting a patch from $x_{k}$ onto $x_{i}$: 
$m_x(x_{i}, x_{k}, \lambda) = \mathbb{1}_m \odot x_{i} +
(\mathbb{1}-\mathbb{1}_m) \odot x_{k}$ where $\odot$ represents the element-wise
product and $\mathbb{1}_m$ a binary mask with average
value $\lambda$. CutMix randomly samples squares, which often leads to rectangular masks due to boundary effects. Such
\textbf{non-linear binary masking} improves generalization
\cite{summers2019improved,takahashi2019data} by increasing dataset: it creates new images with usually disjoint patches
\cite{harris2020mix}.
\cite{BAEK2021107594,faramarzi2020patchup} seek more diverse transformations
via arbitrarily shaped masks: proposals range from
cow-spotted masks \cite{french2020milking} to masks with irregular
edges \cite{harris2020mix}.
As masking of discriminative regions may cause label
misallocation \cite{guo2019mixup},
\cite{kim2020puzzle,uddin2020saliencymix}
try to alleviate this issue with costly saliency heatmaps
\cite{selvaraju2017grad}. Yet, ResizeMix \cite{qin2020resizemix} shows that they
perform no better than random selection of patch locations.

In addition to Manifold Mixup \cite{manifoldmixup19}, only a few works \cite{faramarzi2020patchup,li2020n,yaguchi2019mixfeat,yun2019cutmix} have
tried to mix intermediate \textbf{latent features} as we do.
Our goals and methods are however quite different, as shown later in Section \ref{sec:msdamixmo}.
In brief, they mix deep features to smooth the decision boundaries, while we mix shallow features only so that inputs can remain distinct.

\subsection{Ensembling}%

\paragraph{}%
Like \cite{wen2021combining}, we explore combining DA with another standard technique in machine learning: ensembling \cite{dietterich2000ensemble,hansen1990neural}. For improved performances, aggregated members should be both \textit{accurate} and \textit{diverse} \cite{opitz1999popular,perrone1992networks,rame2021dice}.
Deep ensembles \cite{lakshminarayanan2016simple} (DE) simultaneously train multiple networks with different random initializations converging towards different explanations for the training data \cite{fort2019deep,wilson2020bayesian}.

Ensembling's fundamental drawback is the inherent \textbf{computational and memory overhead}, which
increases linearly with the number of members. This bottleneck is typically addressed by sacrificing either \textit{individual} performance or \textit{diversity} in a complex \textit{trade-off}. Averaging predictions from several checkpoints on the training process, \textit{i.e.}\ snapshot ensembles \cite{huang2017snapshot,izmailov2018averaging}, fails to explore multiple local optima \cite{ashukha2020pitfalls,fort2019deep,wilson2020bayesian}. So does Monte Carlo Dropout \cite{gal2016dropout}. The recent BatchEnsemble
\cite{dusenberry2020efficient}
is parameter-efficient, yet requires multiple forward passes.
TreeNets \cite{lee2015m,szegedy2015going} reduce training and inference cost by sharing low-level layers. MotherNets \cite{wasay2018othernets} share first training epochs between members.
However, sharing reduces diversity.


Very recently, the multi-input multi-output MIMO \cite{havasi2020raining} achieves \textbf{ensemble almost ``for free''}: all of the layers except the first convolutional and last dense layers are shared ($\approx+1\%$ \#parameters). \cite{soflaei2020aggregated} motivated a related Aggregated Learning to learn concise representations with arguments from information bottleneck \cite{tishby2000information}. The idea is that over-parameterized CNNs \cite{lottery2019,molchanov2016pruning,pensia2020ptimal} can fit multiple subnetworks \cite{veit2016residual}. The question is how to prevent homogenization among the simultaneously trained subnetworks.
Facing a similar challenge, \cite{gao2019ntraensemble} includes stochastic channel
recombination; \cite{durasov2020asksembles} relies on predefined binary masks; in GradAug \cite{yang2020radaug}, subnetworks only leverage the first channels up to a given percentage. In contrast, MIMO does not need structural differences among subnetworks: they learn to build their own paths while being as diverse as in DE.%

%% file: sections/03_model.tex
\section{MixMo framework}
We first introduce the main components of our MixMo strategy, summarized in Fig.~\ref{fig:archi_network}: we mix multiple inputs to obtain multiple outputs via subnetworks.
We highlight the key mixing block combining information from inputs, and our training loss based on a dedicated weighting scheme.%

We mainly study $M=2$ subnetworks here, both for clarity and as it empirically
performs best in standard parameterization regimes. For completeness, we
straightforwardly generalize to $M>2$ in Section \ref{model:mheads}.%

\subsection{General overview}
We leverage a training classification dataset $D$ of i.i.d. pairs of associated image/label $\{x_i, y_{i}\}_{i=1}^{|D|}$. We randomly sample a subset of $|B|$ samples $\{x_i, y_{i}\}_{i \in B}$ that we randomly shuffle via permutation $\pi$. Our training batch is $\{(x_i, x_{j}), (y_{i}, y_{j})\}_{i \in B, j=\pi(i)}$. The loss $\mathcal{L}_{\text{MixMo}}$ is averaged over these $|B|$ samples: the networks' weights are updated through backpropagation and gradient descent.%

Let's focus on the training sample $\{(x_0, x_1), (y_0, y_1)\}$. In MixMo, both inputs are \textbf{separately encoded} (see Fig.~\ref{fig:intro}) into the shared latent space with two different convolutional layers (with $3$ input channels each and no bias term): $x_0$ via $c_0$ and $x_1$ via $c_1$. To recover a strictly equivalent formulation to MIMO \cite{havasi2020raining}, we simply sum the two encodings: $c_{0}(x_0) + c_{1}(x_1)$. Indeed, MIMO merges inputs through channel-wise concatenation in pixels: MIMO's first convolutional layer (with $6$ input channels and no bias term) hides the summing operation in the output channels.%

Explicitly highlighting the underlying mixing leads us to consider a \textbf{generalized multi-input mixing block} $\mathcal{M}$. This manifold mixing presents a unique opportunity to tackle the ensemble diversity/individual accuracy trade-off and to improve overall ensemble results (see Section \ref{section:fusion}). The shared representation $\mathcal{M}\left(c_0(x_0), c_1(x_1)\right)$ feeds the next convolutional layers. We note $\kappa$ the \textbf{mixing ratio} between inputs.%

The core network $\mathcal{C}$ handles features that represent both inputs simultaneously. The dense layer $d_{0}$ predicts $\hat{y}_{0} = d_{0}\left[\mathcal{C}\left( \mathcal{M} \left\{c_0(x_0), c_{1}(x_1)\right\} \right)\right]$ and targets $y_{0}$, while $d_{1}$ targets $y_1$. Thus, the \textbf{training loss} is the sum of two cross-entropies $\mathcal{L}_{\text{CE}}$ weighted by parametrized function $w_{r}$ (defined in Section \ref{section:weight}) to balance the asymmetry when $\kappa \ne 0.5$:%
\begin{equation}%
\label{eq:mainloss}%
\mathcal{L}_{\text{MixMo}}=w_{r}(\kappa)\mathcal{L}_{\scalemath{0.59}{\text{CE}}}\left(y_{0},\hat{y}_{0}\right)+w_{r}(1\scalemath{1.}{-}\kappa)\mathcal{L}_{\scalemath{0.59}{\text{CE}}}\left(y_{1}, \hat{y}_{1}\right).%
\end{equation}%
At inference, the same input $x$ is repeated twice: the core network $\mathcal{C}$ is fed the sum $c_{0}(x) + c_{1}(x)$ that preserves maximum information from both encodings. Then, the diverse predictions are averaged: $\frac{1}{2}\left( \hat{y}_{0} + \hat{y}_{1} \right)$. This allows us to benefit from ensembling in a single forward pass.%

\subsection{Mixing block $\mathcal{M}$}
\label{section:fusion}
The mixing block $\mathcal{M}$ - which combines both inputs into a shared representation - is the cornerstone of MixMo.
%
Our main intuition was to analyze MIMO as a simplified Mixup variant where the mixing ratio $\kappa$ is fixed to $0.5$.
MixMo generalized framework encompasses a wider range of variants inspired by MSDA mixing methods. 
Our first main variant - Linear-MixMo - fully extends Mixup. The mixing block is $\scalemath{0.9}{\mathcal{M}_{\text{Linear-MixMo}}}\left(l_{0}, l_{1}\right) =
2\left[\kappa l_{0} + (1-\kappa) l_{1}\right]$, where $l_{0}=c_{0}(x_0)$, $l_{1}=c_{1}(x_1)$ and $\kappa \sim \text{Beta}(\alpha,\alpha)$ with $\alpha$ the concentration parameter. The second and more effective variant \textbf{Cut-MixMo} adapts the patch mixing from CutMix:%
\begin{equation}%
\mathcal{M}_{\text{Cut-MixMo}}\left(l_{0}, l_{1}\right) = 2\left[\mathbb{1}_{\mathcal{M}}\scalemath{1.0}{\odot}l_{0}+(\mathbb{1}-\mathbb{1}_{\mathcal{M}})\scalemath{1.0}{\odot} l_{1}\right],\label{eq:cutmixmo}%
\end{equation}%
\noindent%
where $\mathbb{1}_{\mathcal{M}}$ is a binary mask with area ratio $\kappa \sim \text{Beta}(\alpha,\alpha)$, valued at $1$ either on a rectangle or on the complementary of a rectangle.
In brief, a patch from $c_{0}(x_0)$ is pasted onto $c_{1}(x_1)$, or vice versa.
This binary mixing in Cut-MixMo advantageously replaces the linear interpolation in MIMO and Linear-MixMo: subnetworks are more accurate and more diverse, as shown empirically in Fig.~\ref{fig:tradeoff_probcutmix}.

First, binary mixing in $\mathcal{M}$ trains stronger \textbf{individual}
subnetworks for the same reasons why CutMix improves over Mixup.
In a nutshell, linear MSDAs \cite{manifoldmixup19,zhang2018mixup} produce noisy samples \cite{onmixup2020} that lead to robust representations.
As MixMo tends to distribute different inputs on
non-overlapping channels (as discussed later in Fig.~\ref{fig:mixmo_proj}), this regularization hardly takes place anymore in
$\mathcal{M}_{\text{Linear-MixMo}}$.
On the contrary, by masking features,
we simulate common object occlusion problems. This spreads
subnetworks' focus across different locations: the two classifiers are forced to
find information relevant to their assigned input at disjoint locations. This
occlusion remains effective as the receptive field in this first shallow latent
space remains small.%

Secondly, linear interpolation is fundamentally ill-suited to induce
diversity as
full information is preserved from both inputs.
CutMix on the other hand explicitly increases dataset diversity by presenting
patches of images that do not normally appear together.
Such benefits can be directly transposed to $\mathcal{M}_{\text{Cut-MixMo}}$:
binary mixing with patches increases randomness and \textbf{diversity between the subnetworks}.
Indeed, in a similar spirit to bagging
\cite{breiman1996bagging}, different samples are given to the subnetworks. By
deleting asymmetrical complementary locations from the two inputs, subnetworks
will not rely on the same region and information. Overall, they are less likely
to collapse on close solutions.

\input{pictures/model/archi_network.tex}%
\subsection{Loss weighting $w_r$}%
\label{section:weight}
\input{sections/03_model_reweight_shared.tex}%
\input{pictures/experiments/feat_model.tex}%
\subsection{From manifold mixing to MixMo}
\label{sec:msdamixmo}
We have discussed at length how we extend multi-input multi-output frameworks by
borrowing mixing protocols from MSDA. Now we reversely point out how
our \textbf{MixMo diverges from MSDA schemes}. At first glimpse, the idea is the
same as manifold mixing
\cite{faramarzi2020patchup,li2020n,manifoldmixup19}: $M=2$ inputs
are encoded into a latent space to be mixed before being fed to the rest of the network.
Yet, while they mix at varying depths, we only mix in the shallowest space. Specifically, we only mix in features - and not in pixels - to allow separate encodings of the inputs: they need to remain distinct in the mixed representation for the subsequent classifiers.

Hence our two key differences: \textit{first}, MixMo uses two separated encoders (one for each input), and \textit{second}, it outputs two predictions instead of a single one. Indeed, MSDAs use a single classifier that targets a unique soft label reflecting the different classes via linear interpolation. MixMo instead chooses to fully leverage the composite nature of mixed samples and trains
\textbf{separated dense layers}, $d_{0}$ and $d_{1}$, ensembled ``for free'' at test
time.

Section \ref{expe:differentencdec} demonstrates that MixMo works because it also
uses two \textbf{different encoders} $c_0$ and $c_1$.
While training two classifiers may seem straightforward in MSDA, it actually
raises a troubling question: which input should each classifier predicts ?
Having two encoders provides a simple solution: the network is divided in two subnetworks, one for each input. Their separability is easily observed:
Fig.~\ref{fig:mixmo_proj} shows the $l_1$-norm of the 16 filters
for the two encoders (WRN-28-10 on CIFAR-100). Each filter norm is far from
zero in only one of the two encoders: $c_0(x_0)$ and $c_1(x_1)$ separate the inputs in different dimensions which allows subsequent layers to treat them differently.

This leads MixMo to use most available filters. Following the
structured pruning literature \cite{li2017pruning}, we consider in
Fig.~\ref{fig:active_feats} that a filter (in a layer of the core network) is
active if its $l_1$-norm is at least $40\%$ of the $l_1$-norm from its layer's
most active filter (see Appendix \ref{app:feature}).
This illustrates the known increase in sparsity in wider networks. Conversely, having 2 subnetworks in MixMo enables the weights ignored by one subnetwork to be leveraged by the other.
\subsection{Generalization to $M\geq2$ subnetworks}%
\label{model:mheads}%
Most of the framework is easily extended by optimizing \linebreak 
$\mathcal{L}_{\text{MixMo}}=\sum_{0 \leq i < M}
M\frac{\kappa_i^{1/r}}{\sum_{j}\kappa_j^{1/r}}\mathcal{L}_{\scalemath{0.59}{\text{CE}}}\left(y_{i},\hat{y}_{i}\right)$
with $\{\kappa_i\} \sim \text{Dir}(\alpha)$ from a Dirichlet distribution (see Appendix \ref{app:mheads}). The key change is that $\mathcal{M}$ now needs to
handle more than 2 inputs: $\{c_i(x_i)\}_{0 \leq i < M}$. While linear
interpolation is easily generalized, Cut-MixMo has several possible extensions:
in our experiments, we first linearly interpolate between $M-1$ inputs and then
patch in a region from the $M$-th.%

%% file: pictures/model/archi_network.tex
\begin{figure}%
\includegraphics[width=\linewidth]{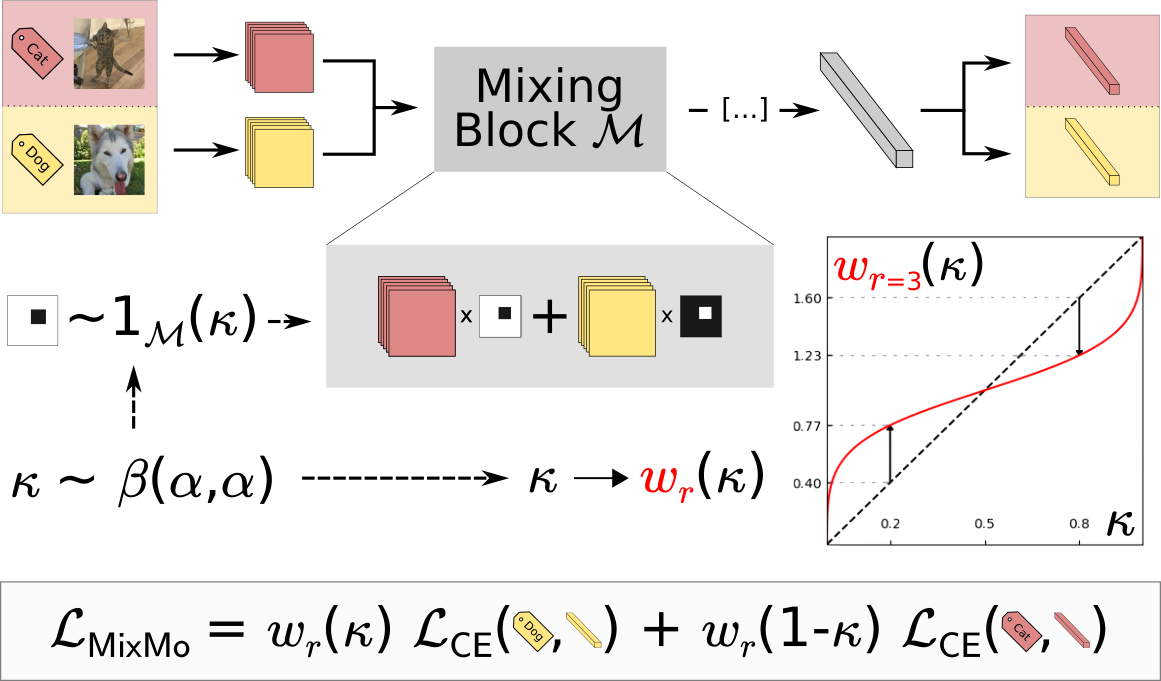}%
\caption{\textbf{Cut-MixMo training}. We sample a mixing mask given $\kappa$, and balance the losses with $w_{r}(\kappa)$ from Eq.~\ref{eq:wrkappa}.}%
\label{fig:archi_network}%
\end{figure}%

%% file: sections/03_model_reweight_shared.tex
Asymmetries in the mixing mechanism can cause one
input to overshadow the other. Notably when $\kappa \ne 0.5$, the predominant input may be easier to predict.
We seek a weighting function $w_r$ to \textbf{balance the relative importance} of the two $\mathcal{L}_{\text{CE}}$ in
$\mathcal{L}_{\text{MixMo}}$. This weighting modifies the effective learning rate, how gradients flow in the network and overall how mixed information is represented in features.
In this paper, we propose to weight via the parametrized:%
\begin{equation}%
w_{r}(\kappa) = 2\frac{\kappa^{1/r}}{\kappa^{1/r} + (1-\kappa)^{1/r}}.
\label{eq:wrkappa}
\end{equation}%
This defines a family of functions indexed by the parameter $r$, visualized for $r=3$ in red on Fig.~\ref{fig:archi_network}. See Appendix \ref{app:weighting} for complementary visualizations. This power law provides a natural relaxation between \textbf{two extreme configurations}. The first extreme, $r=1$, $w_{1}(\kappa)=2\kappa$, is in line with linear label interpolation in MSDA. The resulting imbalance in each subnetwork's contribution to $\mathcal{L}_{\text{MixMo}}$ causes lopsided updates. While it promotes diversity, it also reduces regularization: the overshadowed input has a reduced impact on the loss. The opposite extreme, $r\to\infty$, $w_{\infty}(\kappa)\to1$, removes reweighting. Consequently, $w_r$ inflates the importance of hard under-represented inputs, \textit{à la} Focal Loss \cite{lin2017focal}. However, minimizing the role of the predominant inputs destabilizes training. Overall, we empirically observe that moderate values of $r$ perform best as they trade off pros and cons from both extremes.%

Interestingly, the proper weighting of loss components is also a central theme in \textbf{multi-task learning} \cite{caruana1997multitask,chen2018gradnorm}. While it aims at predicting several
tasks from a shared input, MixMo predicts a shared task from several different inputs.
Beyond this inverted structure, we have similar issues: \textit{e.g.}\ gradients for one task can be detrimental to another conflicting task. Fortunately, MixMo presents an advantage: the exact ratios $\kappa$ and $1-\kappa$ of each task are known exactly.%

%% file: pictures/experiments/feat_model.tex
\begin{figure}
\centering
\begin{subfigure}[b]{0.48\columnwidth}
\centering
\includegraphics[width=\columnwidth]{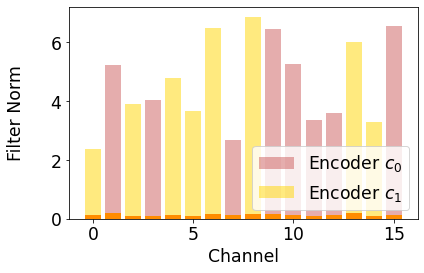}
\caption{Filters $l_1$-norms of the input encoders $c_0$ and $c_1$.}
\label{fig:mixmo_proj}
\end{subfigure}
\hfill%
\begin{subfigure}[b]{0.48\columnwidth}%
\centering%
\includegraphics[width=\columnwidth]{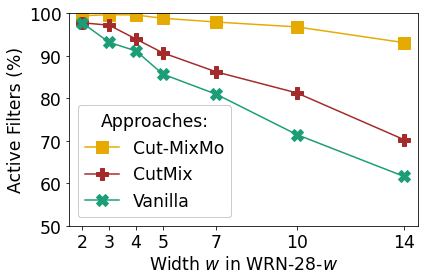}%
\caption{Proportion of active filters in the core network vs.\ width $w$.}%
\label{fig:active_feats}%
\end{subfigure}%
\caption{\textbf{Influence of MixMo on network utilization}.\linebreak(a) The encoders have separate channels: the two subsequent classifiers can differentiate the two inputs. (b) Less filters are strongly active ($\lVert f_i \rVert_1 \geq 0.4\scalemath{1.}{\times}\max_{f\in \text{layer}} \lVert f \rVert_1$) in wider networks: Cut-MixMo reduces this negative point.}
\label{fig:proj_model}
\vspace{-0.5em}%
\end{figure}

%% file: sections/04_experiments.tex
\section{Experiments}
\label{expe:experiments}%
We evaluate MixMo efficiency on standard image classification datasets: CIFAR-\{10,100\} \cite{krizhevsky2009learning} and Tiny ImageNet \cite{chrabaszcz2017}. We equally track accuracies (Top\{1,5\}, $\uparrow$) and the \textit{calibrated Negative Log-Likelihood} (NLL$_{c}$,
$\downarrow$). Indeed, \cite{ashukha2020pitfalls} shows that we should compare in-domain uncertainty estimations after temperature scaling (TS) \cite{guo2017calibration}: we thus split the test set in two and calibrate (after averaging in ensembles) with the temperature optimized on the other half, as in \cite{lobacheva2020power,rame2021dice}. We nonetheless report NLL (without TS) along with the Expected Calibration Error \cite{naeini2015obtaining} in Appendix \ref{app:expeeval}.%
\input{sections/03_xdetails.tex}%
\subsection{Main results on CIFAR-100 and CIFAR-10}%
\input{tables/main_cifar10100}%
Tab.~\ref{table:cifar10010} reports averaged scores over 3 runs for our main experiment on CIFAR with WRN-28-10 \cite{BMVC2016_87}. We re-use the hyper-parameters given in MIMO \cite{havasi2020raining}. Cut-MixMo reaches ($85.40\%$ Top1, $0.535$ NLL$_c$) on
CIFAR-100 with $b\scalemath{1.}{=}4$: it surpasses our Linear-MixMo ($83.08\%$, $0.656$) and MIMO ($83.06\%$, $0.661$). Cut-MixMo sets a new state of the art when combined with CutMix ($85.77\%$, $0.524$).
Results remain strong when $b\scalemath{1.}{=}2$: Cut-MixMo ($84.38\%$, $0.563$) proves better on its own than traditional DE \cite{lakshminarayanan2016simple}, and MSDAs like MixUps \cite{zhang2018mixup, manifoldmixup19} or the stronger CutMix variant \cite{yun2019cutmix}.
On CIFAR-10, we see similar trends: Cut-MixMo reaches $0.081$ in NLL$_c$, $0.079$ with CutMix. Yet, the costlier batch augmented Mixup BA \cite{Hoffer_2020_CVPR} edges it out in Top1.%

Fig.~\ref{fig:paramefficient} shows how MixMo grows stronger than DE (green curves) as width $w$ in WRN-28-$w$ increases. The parameterization becomes
appropriate at $w\scalemath{1.}{=}4$:
Cut-MixMo (yellow curves) then matches DE - with half the parameters - in
Fig.~\ref{fig:widthcx} and its subnetworks match a vanilla
network in Fig.~\ref{fig:indacc}. Beyond, MixMo better uses over-parameterization:
Cut-MixMo+CutMix surpasses DE+CutMix in NLL$_c$ for $w\scalemath{1.}{\geq}5$, and this is true in Top1 for $w\scalemath{1.}{\geq}10$.
Compared to our strong Linear-MixMo+CutMix (purple curves), Cut-MixMo performs similarly in Top1, and better with CutMix for $w\scalemath{1.}{\geq}4$.
While Linear-MixMo and DE learn from occlusion, Cut-MixMo also benefits from CutMix, notably from the induced label smoothing.
Overall, Cut-MixMo, even without CutMix, significantly better estimates uncertainty.%
\input{pictures/experiments/fig_acc_ind.tex}%
\subsection{MixMo analysis on CIFAR-100 w/ WRN-28-10}%
\label{expe:componentanalysis}%
\subsubsection{Training time}%
\label{expe:trainingcost}%
We have just seen that CutMix improves Linear-MixMo at varying widths $w$, but not enough to match Cut-MixMo in NLL$_c$: CutMix can not fully compensate for the advantages from patch mixing over linear interpolation. We recover this finding in Fig.~\ref{fig:batchrepet}, this time at varying batch repetition $b\in\{1,2,4\}$ when $w\scalemath{1.}{=}10$. Moreover, Cut-MixMo outperforms DE for the \textbf{same training time}. Indeed, MixMo variants trained with a given $b$ matches the training time of DE with $N\scalemath{1.}{=}b$ networks. In the rest of this section, we set $b\scalemath{1.}{=}2$.%
\input{pictures/experiments/fig_trainincost.tex}%
\subsubsection{The mixing block $\mathcal{M}$}%
\label{expe:mixingmechanism}%
Tab.~\ref{table:cifar100_othermixingmethod} compares performance for several mixing blocks
\cite{faramarzi2020patchup,harris2020mix,summers2019improved,yun2019cutmix}. No matter the \textbf{shape} (illustrated in Appendix \ref{app:msda}), binary masks perform better than linear mixing: the cow-spotted mask ($84.17\%$, $0.561$) \cite{french2019semi,french2020milking} notably performs well. The basic CutMix patching ($84.38\%$, $0.563$) is nevertheless more accurate and was our main focus.%
\input{tables/table_othermixingmethod.tex}%

We further study the impact of patch mixing through the lens of the
\textit{ensemble diversity/individual accuracy trade off}. As in
\cite{rame2021dice}, we measure diversity via the pairwise ratio-error \cite{aksela2003comparison} ($d_{re}$, $\uparrow$), defined as the ratio between the number of different errors and simultaneous errors for two predictors. In Fig.~\ref{fig:tradeoff_probcutmix} and \ref{fig:tradeoff_root}, we average metrics over the last 10 epochs.%

As argued in Section \ref{section:fusion}, patch mixing increases diversity compared to linear mixing in Fig.~\ref{fig:tradeoff_probcutmix}. As the probability $p$ of patch mixing grows, so does diversity: from $d_{re}(p\scalemath{1.}{=}0.0)\scalemath{1.}{\approx}0.78$ (Linear-MixMo) to $d_{re}(p\scalemath{1.}{=}0.5)\scalemath{1.}{\approx}0.85$ (Cut-MixMo). We provide associated training dynamics in Appendix \ref{app:td}. In contrast, DE has $d_{re}\scalemath{1.}{\approx}0.76$ while MIMO has $d_{re}\scalemath{1.}{\approx}0.77$ on the same setup. Increasing $p$ past $0.6$ boosts diversity even more at the cost of subnetworks' accuracies: this is due to underfitting and an increased test-train
distribution gap. $p\in[0.5, 0.6]$ is thus the best trade off.%
\input{pictures/experiments/fig_tradeoff_probcutmix.tex}%
\subsubsection{Weighting function $w_r$}%
\label{expe:hyperparamablation}
We analyze the impact of the parameter $r$ in the reweighting function $w_r$.
Higher values tend to remove reweighting, as shown in Appendix \ref{app:weighting}: they
strongly decrease diversity in Fig.~\ref{fig:tradeoff_root}.
The opposite extreme with $r\scalemath{1.}{=}1$ increases diversity via lopsided
gradient updates but it degrades accuracy.
We speculate it under-emphasizes hard samples.
The range $r\in[3,6]$ strikes a good balance: results remain high and stable.%
\input{pictures/experiments/fig_tradeoff_root.tex}%
\subsubsection{Generalization to $M\geq2$ subnetworks}%
\label{expe:mheads}%
We try to generalize MixMo to more than $M=2$ subnetworks in Fig.~\ref{fig:msubnetworks}. Cut-MixMo's subnetworks perform at $82.3\%$
when $M\scalemath{1.}{=}2$ vs.\ $79.5\%$ when $M\scalemath{1.}{=}3$. In MIMO,
it's $79.8\%$ vs.\ $77.7\%$. Because subnetworks do not share features, higher
$M$ degrades their results: only two can fit seamlessly.
Ensemble Top1 overall decreases in spite of the additional predictions, as already noticed in MIMO \cite{havasi2020raining}.%
\newpage
\input{pictures/experiments/fig_msubnetworks_w}%
This reflects MixMo's strength in
over-parametrized regimes, but also its limitations with fewer parameters
when subnetworks underfit (recall previous Fig.~\ref{fig:paramefficient}).
Facing similar findings, MIMO \cite{havasi2020raining} introduced input
repetition so that subnetworks share their features, at the cost of drastically
reducing diversity. 
Our generalization may be extended by future approaches whose mixing blocks (perhaps not inspired by MSDA) would tackle these issues.%
\subsubsection{Multiple encoders and classifiers}
\input{tables/ablation_sharedencoders.tex}
\label{expe:differentencdec}
In Section \ref{sec:msdamixmo}, we compared MixMo and MSDA. Tab.~\ref{table:cifar100sh} confirms the need for \textbf{2 encoders and 2 classifiers}. With 1 classifier and linearly interpolated labels (in the same spirit as \cite{chen2020mclr}), the 2 encoders perform worse than 1 encoder. With 1 shared encoder and 2 classifiers, it is not clear which input each classifier should target. In the first naive $\ominus$, we randomly associate the 2 classifiers and the 2 inputs (encoded with the same encoder). This $\ominus$ variant yields poor results. In $\otimes$, the first classifier tries to predict the label from the predominant input, the second targets the other input: $\otimes$ reaches $0.598$ vs.\ $0.563$ for Cut-MixMo.%
\subsection{Robustness to image corruptions}%
Deep networks' results decrease when facing unfamiliar samples. To measure
robustness to train-test distribution gaps, \cite{hendrycks2018benchmarking}
corrupted CIFAR-100 test images into CIFAR-100-c (more details in Appendix \ref{app:expe}).
As in Puzzle-Mix \cite{kim2020puzzle}, we report WRN-28-10 results with and
without AugMix \cite{hendrycks2019augmix}, a pixels data augmentation technique
specifically introduced for this task. Tab.~\ref{table:cifar100c} shows that Cut-MixMo ($b\scalemath{1.}{=}4$) best complements AugMix and reaches $71.1\%$ Top1.%
\input{tables/table_cifar100c_small.tex}%
\subsection{Pushing MixMo further: Tiny ImageNet}%
At a larger scale and with more varied $64\times64$ images, Cut-MixMo reaches a new state of the art of 70.24\% on
Tiny ImageNet \cite{chrabaszcz2017} in Tab.~\ref{table:tiny200}. We re-use the hyper-parameters given in previous state of the art Puzzle-Mix \cite{kim2020puzzle}. With $w\scalemath{1.}{=}1$, PreActResNet-18 \cite{preacthe} is not sufficiently parametrized for MixMo's advantages to express themselves on this challenging dataset. MixMo's full potential shines with wider networks: with $w\scalemath{1.}{=}2$ and $44.9$M parameters, Cut-MixMo reaches ($69.13\%$, $1.28$) vs.\ ($67.76\%$, $1.33$) for CutMix. Compared to DE with $3$ networks,
Cut-MixMo performs \{worse, similarly, better\} for width $w\in\{1,2,3\}$. At (almost) the
same numbers of parameters, Cut-MixMo when $w\scalemath{1.}{=}2$ performs better
($69.13\%$, $1.28$) than DE with 4 networks when $w\scalemath{1.}{=}1$
($67.51\%$, $1.31$).%
\input{tables/main_tiny200.tex}\subsection{Ensemble of MixMo}%
\label{expe:ensemblemixmo}
Since MixMo adds very little parameters ($\approx+1\%$), we can combine independently trained MixMo like in DE. This ensembling of ensemble of subnetworks leads in practice to the
averaging of $M\times N=2\times N$ predictions.
Fig.~\ref{fig:splitadvantage} compares ensembling for vanilla networks
and Cut-MixMo on CIFAR-100. We first recover the Memory Split Advantage
\cite{chirkova2020deep,lobacheva2020power,wang2020ultiple,zhao2020plitnet} (MSA): at similar parameter counts, $N\scalemath{1.}{=}5$ vanilla
WRN-28-3 do better than a single vanilla WRN-28-7 ($+0.10$ in NLL$_c$). \textbf{Cut-MixMo challenges this MSA}: we bridge the gap between using one network or several smaller networks ($-0.04$ on same setup). 
Visually, Cut-MixMo's curves remain closer to the lower envelope: performances are less dependent on how the memory budget is split.
This is because Cut-MixMo is effective mainly for larger architectures by better leveraging their parameters.%
\input{pictures/experiments/fig_sma.tex}%

We also recover that wide vanilla networks tend to be less diverse
\cite{neal2018modern}, and thus gain less from ensembling \cite{lobacheva2020power}: $N\scalemath{1.}{=}2$ vanilla
WRN-28-14 ($83.47\%$ Top1, $0.656$ NLL$_c$) perform not much better than
$N\scalemath{1.}{=}2$ WRN-28-7 ($82.94\%$, $0.673$). Contrarily,
\textbf{Cut-MixMo facilitates the ensembling of large networks} with
($86.58\%$, $0.488$)  vs.\ ($85.50\%$, $0.516$) (more comparisons in Appendix \ref{app:ensmixmo}).

When combined with CutMix \cite{yun2019cutmix}, Cut-MixMo previously set a new state of the art of $85.77\%$ with $N\scalemath{1.}{=}1$ WRN-28-10. Final Tab.~\ref{table:cifar100sota} shows it further reaches $86.63\%$ with $N\scalemath{1.}{=}2$ and even $86.81\%$ with $N\scalemath{1.}{=}3$.
\input{tables/table_sota.tex}

%% file: sections/03_xdetails.tex
\subsection{Implementation details}
We mostly study the Linear-MixMo and Cut-MixMo variants with
$M\scalemath{1.}{=}2$. We set \textbf{hyper-parameter} $r\scalemath{1.}{=}3$ (see Section~\ref{expe:hyperparamablation}). $\alpha\scalemath{1.}{=}2$ performs better than $1$ (see Appendix \ref{app:alpha}). In contrast, MIMO \cite{havasi2020raining} refers to linear summing, like Linear-MixMo, but with
$\kappa\scalemath{1.}{=}0.5$ instead of $\kappa \sim \text{Beta}(\alpha,\alpha)$.

Different mixing methods create a strong \textbf{train-test distribution
gap} \cite{onmixup2020,gontijo2020affinity}. Thus, in Cut-MixMo we actually substitute $\mathcal{M}_{\text{Cut-MixMo}}$ for $\mathcal{M}_{\text{Linear-MixMo}}$
with probability $1-p$ to accommodate for the summing in $\mathcal{M}$ at inference. We set the probability of patch mixing during training to $p\scalemath{1.}{=}0.5$, with linear descent to $0$ over the last twelfth of training epochs (see pseudocode \ref{pseudocode} in Appendix).

When MixMo is combined with CutMix, the pixels inputs are: $\left(m_x(x_{i}, x_{k}, \lambda), m_x(x_{j}, x_{k'}, \lambda')\right)$ with interpolated targets $\left(\lambda y_i +  (1-\lambda)y_{k}, \lambda' y_j +  (1-\lambda')y_{k'})\right)$, where $k,k'$ are randomly sampled and $\lambda,\lambda'\sim\text{Beta}(1,1)$.

MIMO duplicates samples $b$ times via \textbf{batch repetition}:
$x_i$ will be associated with $x_{\pi(i)}$ and $x_{\pi'(i)}$ in the same batch
if $b\scalemath{1.}{=}2$. As the batch size remains fixed, the count of unique
samples per batch and the learning rate is divided by $b$. Conversely, the
number of steps is multiplied by $b$. Overall, this stabilizes training but multiplies
its cost by $b$. We thus indicate an estimated (training/inference) overhead (wrt. vanilla training) in the \textit{time} column of our tables. Note that some concurrent approaches also lengthen training: \textit{e.g.}\ GradAug \cite{yang2020radaug} via multiple subnetworks predictions ($\approx\times3$).

We provide more details in Appendix \ref{app:expe} and will open source our PyTorch \cite{NEURIPS2019_9015} implementation.%

%% file: tables/main_cifar10100.tex
\begin{table}[!t]%
\caption{\textbf{Main results}: WRN-28-10 on CIFAR. \textbf{Bold} highlights best scores, $\dagger$ marks approaches not re-implemented.}%
\centering%
\resizebox{0.86\linewidth}{!}{%
        \begin{tabular}{c | c | c c c | c c}
            \toprule
            \multicolumn{2}{c|}{Dataset} & \multicolumn{3}{c|}{CIFAR-100} & \multicolumn{2}{c}{CIFAR-10} \\
            \midrule
            \midrule
            Approach & \begin{tabular}{@{}c@{}}\scalebox{1.0}{Time}\\\scalebox{1.0}{Tr./Inf.}\end{tabular} & \begin{tabular}{@{}c@{}}\scalebox{1.0}{Top1}\\\scalebox{1.0}{$\%, \uparrow$}\end{tabular} & \begin{tabular}{@{}c@{}}\scalebox{1.0}{Top5}\\\scalebox{1.0}{$\%, \uparrow$}\end{tabular} & \begin{tabular}{@{}c@{}}\scalebox{1.0}{NLL$_{c}$}\\\scalebox{1.0}{$10^{-2}, \downarrow$}\end{tabular} & \begin{tabular}{@{}c@{}}\scalebox{1.0}{Top1}\\\scalebox{1.0}{$\%, \uparrow$}\end{tabular} & \begin{tabular}{@{}c@{}}\scalebox{1.0}{NLL$_{c}$}\\\scalebox{1.0}{$10^{-2}, \downarrow$}\end{tabular} \\
            \midrule
            \midrule
            Vanilla & \multirow{6}{*}{1/1} & 81.63 & 95.49 & 73.9 & 96.34 & 12.6\\
            Mixup & & 83.44 & 95.92 & 65.7 & 97.07 & 11.2 \\
            Manifold Mixup$^\dagger$ & & 81.96 & 95.51 & 73.4 & 97.45 & 12.2 \\
            CutMix & &84.05 & 96.09 & 64.8 & 97.23 & 9.9 \\
            ResizeMix$^\dagger$ & & 84.31 & - & - & 97.60 & -  \\
            \midrule
            Puzzle-Mix$^\dagger$ & 2/1 & 84.31 & 96.46 & 66.8 &-&-\\
            \midrule
            GradAug$^\dagger$ & \multirow{2}{*}{3/1} & 84.14 & 96.43 & - & -&- \\
            + CutMix$^\dagger$ & & 85.51 & 96.86 & - &- & - \\
            \midrule
            Mixup BA$^\dagger$ & 7/1 & 84.30 & - & - & \textbf{97.80} &- \\
            \midrule
            DE (2 Nets) & \multirow{2}{*}{2/2} & 83.17 & 96.37 & 66.4 & 96.67 & 11.1 \\
            + CutMix & & 85.74 & 96.82 & 57.1 & 97.52 & 8.6 \\
            \midrule
            \midrule
            \multirow{1}{*}{MIMO} & \multirow{5}{*}{2/1} & 82.40 & 95.78 & 68.8 & 96.38 & 12.1 \\
            \cmidrule{1-1}
            \cmidrule{3-7}
            \multirow{1}{*}{Linear-MixMo} & & 82.54 & 95.99 & 67.6 & 96.56 & 11.4 \\
            + CutMix & & 84.69 & 97.12 & 57.2 &  97.32 & 9.4 \\
            \cmidrule{1-1}
            \cmidrule{3-7}
            \multirow{1}{*}{Cut-MixMo} & & 84.38 & 96.94 & 56.3 & 97.31 & 8.9 \\
            + CutMix & & 85.18 & 97.20 & 54.5 & 97.45 & 8.4 \\
            \midrule
            \multirow{1}{*}{MIMO} & \multirow{5}{*}{4/1} & 83.06 & 96.23 & 66.1 & 96.74 & 11.4 \\
            \cmidrule{1-1}
            \cmidrule{3-7}
            \multirow{1}{*}{Linear-MixMo} & & 83.08 & 96.26 & 65.6 & 96.91 & 10.8 \\
            + CutMix & & 85.47 & 97.04 & 55.8 & 97.68 & 8.7 \\
            \cmidrule{1-1}
            \cmidrule{3-7}
            \multirow{1}{*}{Cut-MixMo} &  & 85.40 & 97.22 & 53.5 & 97.51 & 8.1 \\
            + CutMix & & \textbf{85.77} & \textbf{97.42} & \textbf{52.4} & 97.73 & \textbf{7.9} \\
            \bottomrule%
        \end{tabular}%
}%
\label{table:cifar10010}%
\vspace{-0.2em}%
\end{table}%

%% file: pictures/experiments/fig_acc_ind.tex
\begin{figure}%
\centering%
\begin{subfigure}{.65\linewidth}%
\centering%
\includegraphics[width=1\linewidth]{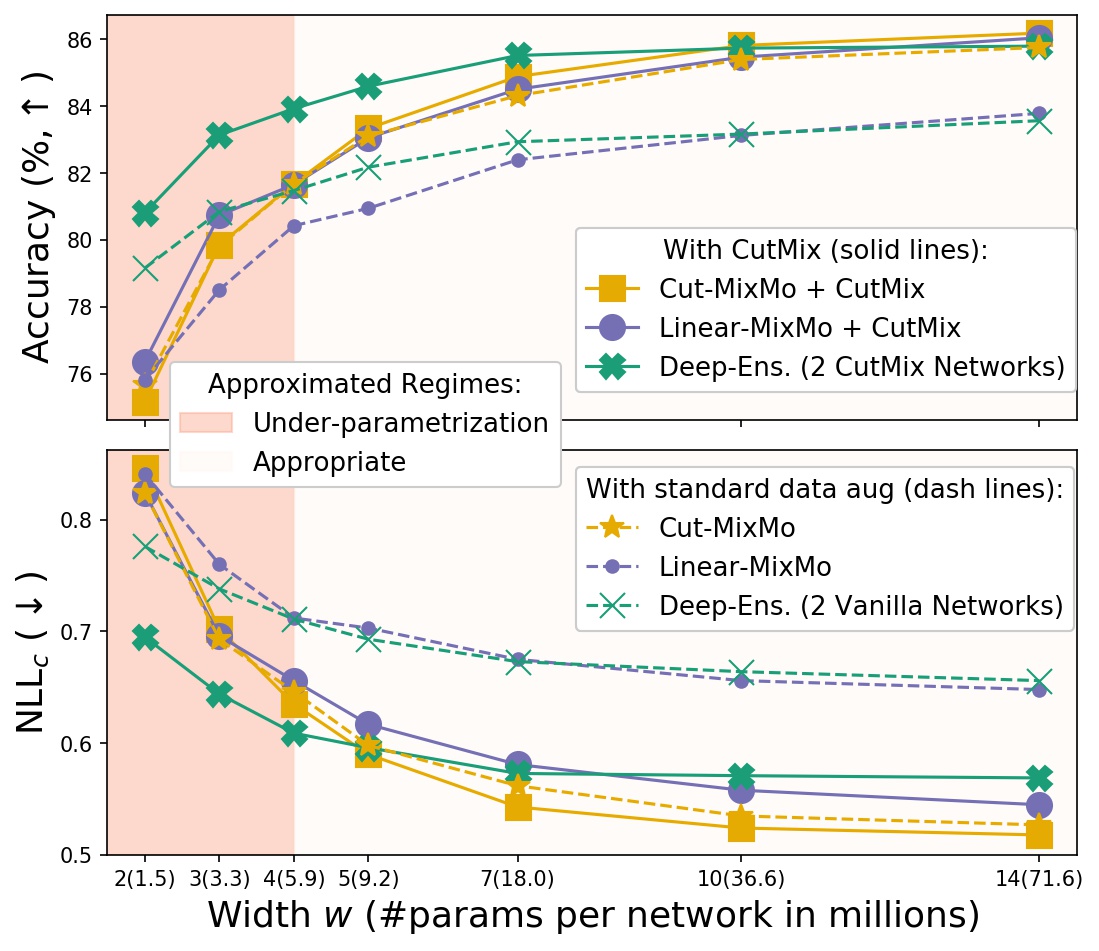}%
\caption{Ensemble Top1 and NLL$_c$.}%
\label{fig:widthcx}%
\end{subfigure}%
\begin{subfigure}{.35\linewidth}%
\centering%
\includegraphics[width=1\linewidth]{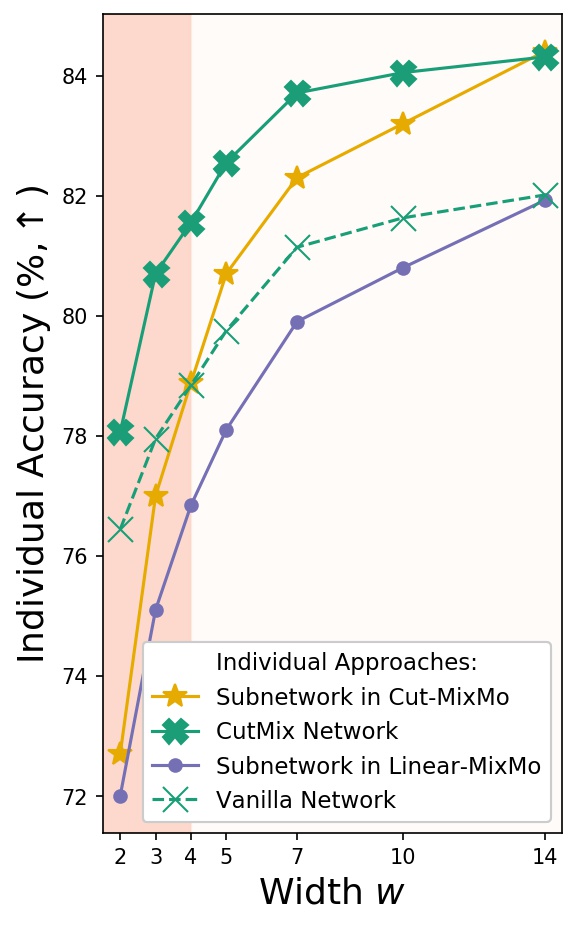}%
\caption{Individual Top1.}%
\label{fig:indacc}%
\end{subfigure}%
\caption{\textbf{Parameters efficiency} (metrics/\#params). CIFAR-100 with WRN-28-$w$, $b\scalemath{1.}{=}4$. Comparisons between (a) ensemble and some of their (b) individual counterparts.}%
\label{fig:paramefficient}%
\end{figure}%

%% file: pictures/experiments/fig_trainincost.tex
\begin{figure}[!h]%
\centering%
\includegraphics[width=0.8\linewidth]{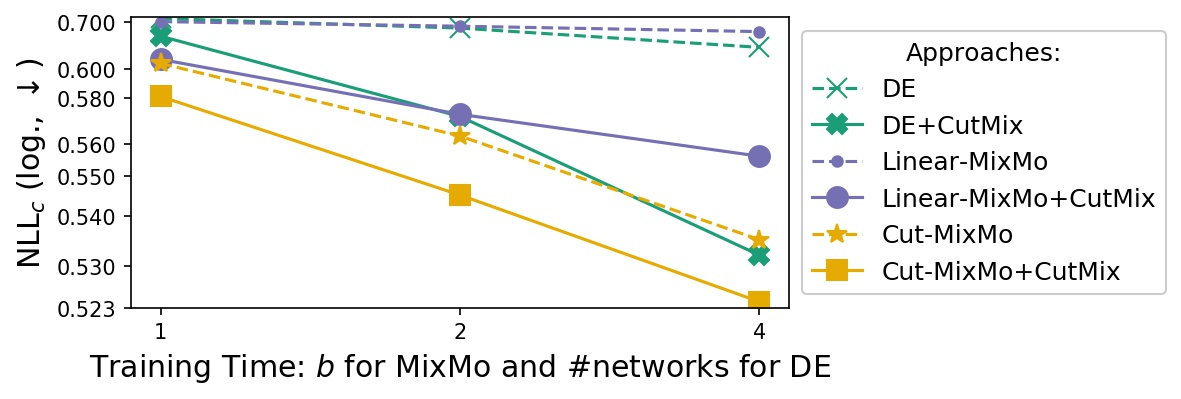}%
\vspace{-0.4em}%
\caption{\textbf{NLL$_c(\downarrow)$ improves with longer training}, via batch repetitions (MixMo) or additional networks (DE).%
}%
\label{fig:batchrepet}%
\vspace{-0.8em}%
\end{figure}%

%% file: tables/table_othermixingmethod.tex
\begin{table}[!h]%
\caption{$\mathcal{M}$ inspired by various \textbf{MSDA approaches}.}%
\centering%
\vspace{-0.4em}%
\resizebox{0.95\linewidth}{!}{%
        \begin{tabular}{c | c | c c c c c c}
            \toprule%
            \begin{tabular}{@{}c@{}}$\mathcal{M}$\\approach\end{tabular} & \begin{tabular}{@{}c@{}}Mixup\\\cite{zhang2018mixup}\end{tabular} & \begin{tabular}{@{}c@{}}Horiz.\\Concat.\end{tabular} & \begin{tabular}{@{}c@{}}Vertical\\Concat.\end{tabular}  & \begin{tabular}{@{}c@{}}PatchUp 2D\\\cite{faramarzi2020patchup}\end{tabular} & \begin{tabular}{@{}c@{}}FMix\\\cite{harris2020mix}\end{tabular} & \begin{tabular}{@{}c@{}}CowMask\\\cite{french2019semi,french2020milking}\end{tabular} & \begin{tabular}{@{}c@{}}CutMix\\\cite{yun2019cutmix}\end{tabular}\\%
            \midrule%
            Top1 $\uparrow$  &82.5 & 82.78 & 84.00 & 84.16 & 83.76 & 84.17 & \textbf{84.38}\\
            NLL$_c$ $\downarrow$ & 0.676 & 0.627 & 0.573 & 0.581 & 0.602 & \textbf{0.561} & 0.563\\
            \bottomrule%
        \end{tabular}}%
\label{table:cifar100_othermixingmethod}%
\end{table}%

%% file: pictures/experiments/fig_tradeoff_probcutmix.tex
\begin{figure}[!h]%
\centering%
\includegraphics[width=0.88\linewidth]{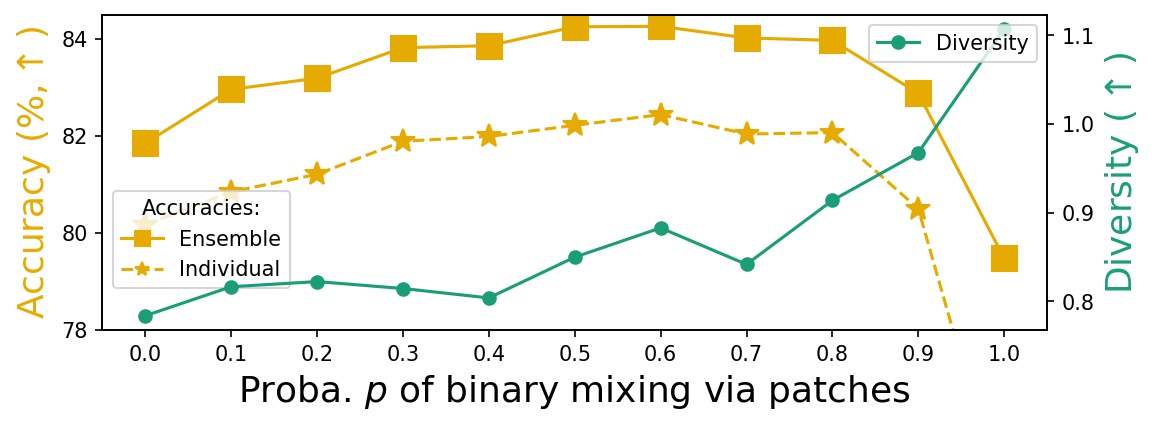}%
\vspace{-0.5em}%
\caption{\textbf{Diversity/accuracy} as function of $p$ with $r\scalemath{1.}{=}3$.}%
\label{fig:tradeoff_probcutmix}%
\vspace{-1em}%
\end{figure}%

%% file: pictures/experiments/fig_tradeoff_root.tex
\begin{figure}[!h]%
\centering%
\includegraphics[width=0.88\linewidth]{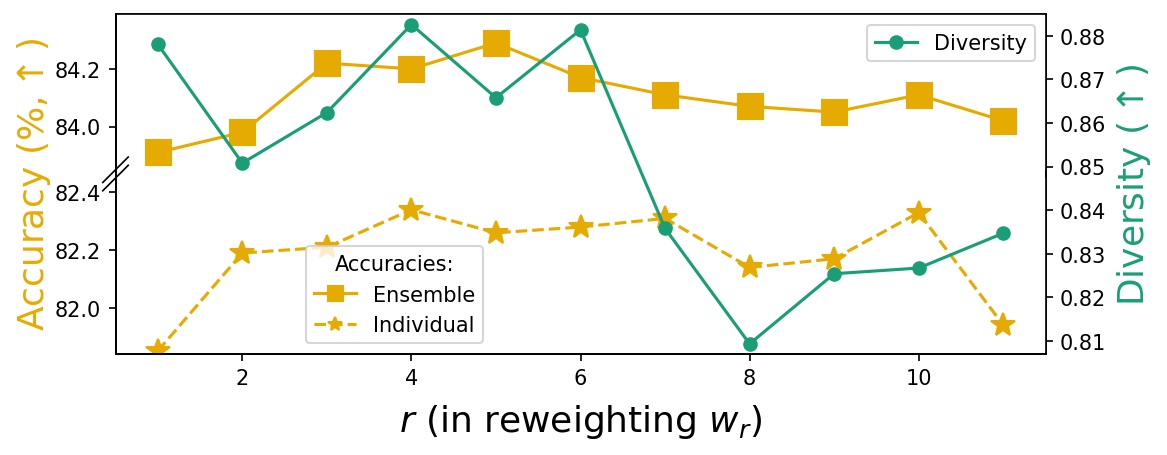}%
\vspace{-0.5em}%
\caption{\textbf{Diversity/accuracy} as function of $r$ with $p\scalemath{1.}{=}0.5$.}%
\vspace{-1em}%
\label{fig:tradeoff_root}%
\end{figure}%

%% file: pictures/experiments/fig_msubnetworks_w.tex
\begin{figure}[!h]%
\centering%
\includegraphics[width=0.77\linewidth]{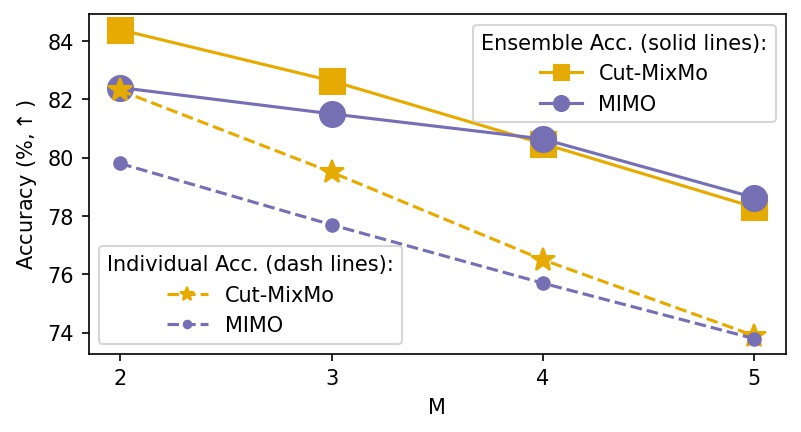}%
\vspace{-1.em}%
\caption{\textbf{Ensemble/individual} accuracies for $M\geq2$.}
\label{fig:msubnetworks}%
\end{figure}%

%% file: tables/ablation_sharedencoders.tex
\begin{wraptable}[10]{hR!}{0.40\linewidth}
\centering
\vspace{-1em}
\caption{\textnormal{Number of encoders/classifiers}.}
\vspace{-0.5em}
\resizebox{\linewidth}{!}{%
        \begin{tabular}{cc |c}
            \toprule
            \# Enc. & \# Clas. & NLL$_c$ $\downarrow$\\
            \midrule
            1 & 1 & 0.604 \\
            2 & 1 & 0.666 \\
            1 & 2$^{\ominus}$ & 0.687 \\
            1 & 2$^{\otimes}$ & 0.598 \\
            \midrule
            2 & 2 & \textbf{0.563} \\
            \bottomrule
        \end{tabular}
}
\label{table:cifar100sh}
\end{wraptable}

%% file: tables/table_cifar100c_small.tex
\begin{table}[!b]%
\vspace{-0.5em}%
\caption{\textbf{Robustness comparison on CIFAR-100-c}.}%
\centering%
\resizebox{1.0\linewidth}{!}{%
        \begin{tabular}{c | c c |c|c c| c c |c |c c |c c}%
            \toprule%
            Approach & \multicolumn{2}{c|}{1 Net.} & CutMix & \multicolumn{2}{c|}{Puzzle-Mix$^\dagger$} & \multicolumn{2}{c|}{DE (2 Nets)} & MIMO & \multicolumn{2}{c|}{Linear-MixMo} & \multicolumn{2}{c}{Cut-MixMo} \\%
            AugMix & - &\checkmark & - & - & \checkmark & - & \checkmark & - & - & \checkmark & - & \checkmark \\%
            \midrule%
            Top1 $\uparrow$ & 52.2 & 67.8 & 51.93 & 58.09 & 70.46 & 53.8 & 69.9 & 53.6 & 55.6 & 70.4 & 57.0 & \textbf{71.1} \\
            Top5 $\uparrow$ & 73.7 & 87.5 & 72.03 & 77.3 & 87.7 & 74.9 & 88.9 &74.9 & 76.1 & 89.4 & 77.4 & \textbf{89.5} \\
            NLL $\downarrow$ & 2.50 & 1.38 & 2.13 & 1.96 & 1.34 & 2.27 & 1.24 & 2.66 & 2.33 & 1.22 & 2.04 & \textbf{1.16} \\%
            \bottomrule%
\end{tabular}}%
\label{table:cifar100c}%
\end{table}%

%% file: tables/main_tiny200.tex
\begin{table}[!h]
\caption{\textbf{Results}: PreActResNet-18-$w$ on Tiny ImageNet.}
\centering
\resizebox{1.0\linewidth}{!}{%
        \begin{tabular}{c | c | c c | c c | c c }
            \toprule
            \multicolumn{2}{c|}{Width $w$ (\# params)} & \multicolumn{2}{c|}{$w=1$ (11.2M)} & \multicolumn{2}{c|}{$w=2$ (44.9M)} & \multicolumn{2}{c}{$w=3$ (100.5M)} \\
            \midrule
            \midrule
            Approach & \begin{tabular}{@{}c@{}}\scalebox{0.9}{Time}\\\scalebox{.8}{Tr./Inf.}\end{tabular} & \begin{tabular}{@{}c@{}}\scalebox{0.9}{Top1}\\\scalebox{0.8}{$\%, \uparrow$}\end{tabular} & \begin{tabular}{@{}c@{}}\scalebox{0.9}{NLL$_{c}$}\\\scalebox{0.8}{$\downarrow$}\end{tabular} &
           \begin{tabular}{@{}c@{}}\scalebox{0.9}{Top1}\\\scalebox{0.8}{$\%, \uparrow$}\end{tabular} & \begin{tabular}{@{}c@{}}\scalebox{0.9}{NLL$_{c}$}\\\scalebox{0.8}{$\downarrow$}\end{tabular} &
           \begin{tabular}{@{}c@{}}\scalebox{0.9}{Top1}\\\scalebox{0.8}{$\%, \uparrow$}\end{tabular} & \begin{tabular}{@{}c@{}}\scalebox{0.9}{NLL$_{c}$}\\\scalebox{0.8}{$\downarrow$}\end{tabular}
            \\
            \midrule
            \midrule
            Vanilla & \multirow{5}{*}{1/1} & 62.56 & 1.53 & 64.80 & 1.51 & 65.78 & 1.53\\
            Mixup & & 63.74 & 1.62 & 66.62 & 1.50 & 67.27 & 1.51\\
            Manifold Mixup$^{\dagger}$ &  & 58.70 & 1.92 & -&- &- &- \\
            Co-Mixup$^{\dagger}$ & & 64.15 & - & - & - &- &- \\
            CutMix & & 65.09 & 1.58 & 67.76 & 1.33 & 68.95 & 1.29 \\
            \midrule
            Puzzle-Mix$^{\dagger}$ & 2/1 & 64.48 & 1.65 &-&- &- &- \\
            \midrule
            DE (2 Nets) & 2/2 & 65.53 & 1.39 & 68.06 & 1.37 & 68.38 & 1.36 \\
            DE (3 Nets) & 3/3 & 66.76 & 1.34 & 69.05 & 1.29 & 69.36 & 1.28 \\
            DE (4 Nets) & 4/4 & \textbf{67.51} & \textbf{1.31} & \textbf{69.94} & \textbf{1.24} & 69.72 & 1.26 \\
            \midrule
            \midrule
            \multirow{1}{*}{Linear-MixMo} & \multirow{2}{*}{2/1} & 61.58 & 1.61 & 66.62 & 1.41 & 68.18 & 1.36 \\
            \multirow{1}{*}{Cut-MixMo} & & 63.78 & 1.48 & 68.30 & 1.30 & 69.89 & 1.26 \\
            \midrule
            \multirow{1}{*}{Linear-MixMo} & \multirow{2}{*}{4/1} & 62.91 & 1.51 & 67.03 & 1.41 & 68.38 & 1.38\\
            \multirow{1}{*}{Cut-MixMo} & & 64.44 & 1.48 & 69.13 & 1.28 & \textbf{70.24} & \textbf{1.19}\\
            \bottomrule%
\end{tabular}}
\label{table:tiny200}
\end{table}

%% file: pictures/experiments/fig_sma.tex
\begin{figure}[!t]%
\centering%
\includegraphics[width=1.0\linewidth]{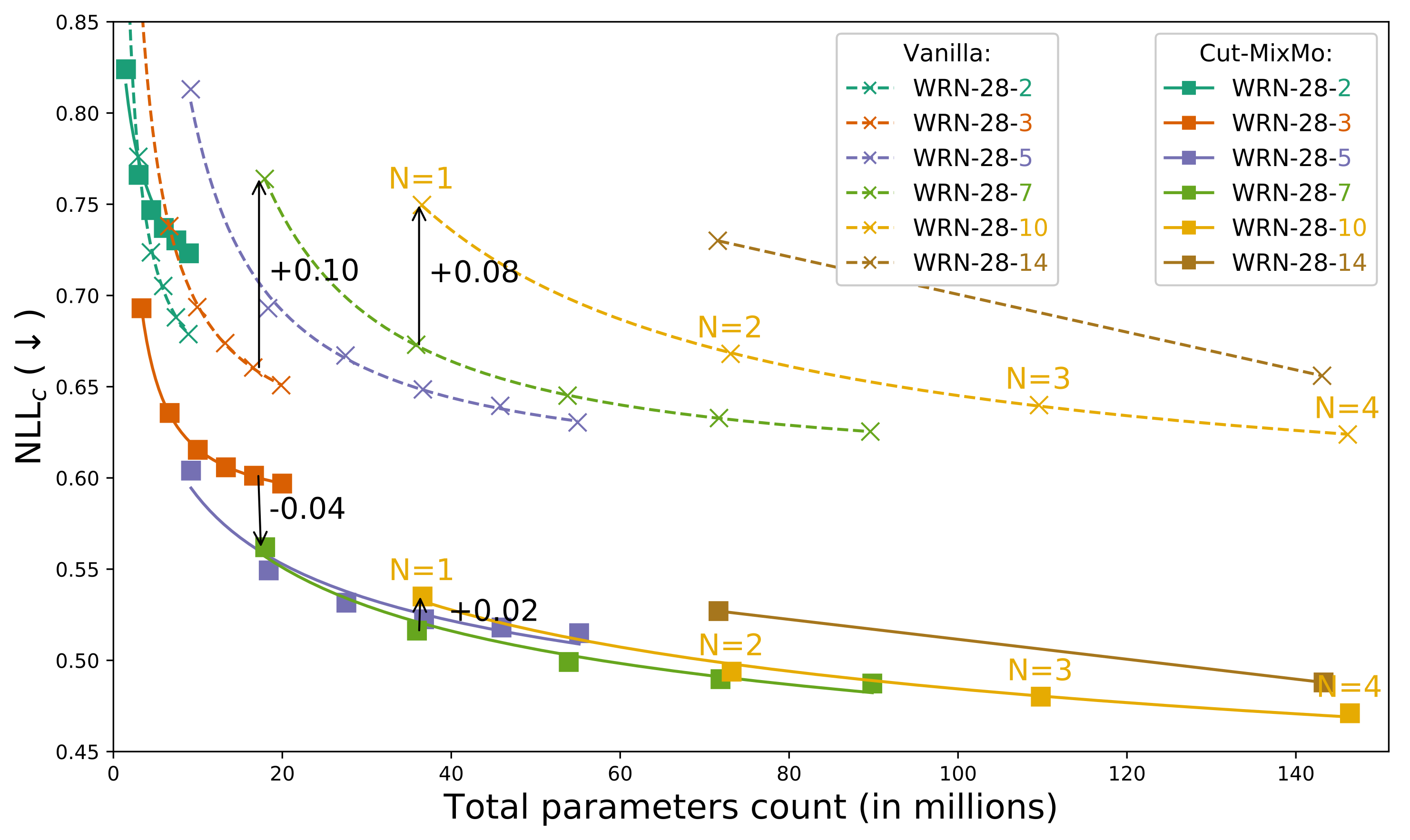}%
\caption{\textbf{Ensemble effectiveness} (NLL$_c$/\#params), for different widths $w$ in WRN-28-$w$ and numbers of members $N$. Standard data augmentations on CIFAR-100 with $b\scalemath{1.}{=}4$. Curves interpolated through power laws \cite{lobacheva2020power}.}%
\label{fig:splitadvantage}%
\end{figure}%

%% file: tables/table_sota.tex
\begin{table}[H]
\caption{\textbf{Best results for WRN-28-10 on CIFAR-100} via Cut-MixMo + CutMix \cite{yun2019cutmix} + $N$-ensembling and $b\scalemath{1.}{=}4$.\linebreak Recent Top1 SoTAs: 85.23 \cite{qin2020resizemix}, 85.51 \cite{yang2020radaug}, 85.74 \cite{zhao2020plitnet}.}%
\centering
\resizebox{0.49\textwidth}{!}{%
        \begin{tabular}{c  c | c c c| c c c}
            \toprule
            \multirow{2}{*}{$N$} & \multirow{2}{*}{\# params} & \multicolumn{3}{c|}{Average} & \multicolumn{3}{c}{Best run} \\
            & & Top1 $\uparrow$ & Top5 $\uparrow$ & NLL$_c$ $\downarrow$ & Top1 $\uparrow$ & Top5 $\uparrow$ & NLL$_c$ $\downarrow$\\
            \midrule
            1 & 36.6M & 85.77 \scalebox{.7}{$\pm$ 0.14} & 97.36 \scalebox{.7}{$\pm$ 0.02} & 0.524 \scalebox{.7}{$\pm$ 0.005} & 85.92 & 97.36 & 0.518\\
            2 & 73.2M &  86.63 \scalebox{.7}{$\pm$ 0.19} & 97.73 \scalebox{.7}{$\pm$ 0.05} & 0.479 \scalebox{.7}{$\pm$ 0.003} & 86.75 & 97.80 & 0.475\\
            3 & 109.8M & 86.81 \scalebox{.7}{$\pm$ 0.17} & 97.85 \scalebox{.7}{$\pm$ 0.04} & 0.464 \scalebox{.7}{$\pm$ 0.002} & 86.94 & 97.83 & 0.464\\
            \bottomrule%
\end{tabular}%
}
\label{table:cifar100sota}
\end{table}

%% file: sections/05_conclusion.tex
\section{Conclusion}%
We introduce the MixMo framework that generalizes the multi-input multi-output ensembling strategy.
MixMo can be analyzed as either an ensembling method or a mixed samples data
augmentation, while remaining complementary to works from both lines of
research.
Finally, MixMo better exploits wide networks and improves the state of the art on CIFAR-100, CIFAR-100-c and Tiny ImageNet.

%% file: sections/06_appendix.tex
The sections in this Appendix follow a similar order to their related sections in the main paper. 
We first illustrate the reweighting of the loss components in Appendix \ref{app:weighting}.
Appendix \ref{app:feature} elaborates on our analysis of filters activity.
Appendix \ref{app:mheads} clarifies our framework generalization with $M>2$ subnetworks.
We describe in greater details our implementation in Appendix \ref{app:expe}, and then our evaluation setting in \ref{app:expeeval}.
Appendix \ref{app:td} showcases training dynamics.
We provide a quick refresher on common MSDA techniques in Appendix \ref{app:msda}.
Appendix \ref{app:alpha} studies the importance of $\alpha$.
Appendix \ref{app:imagenet} is a preliminary study of MixMo on ImageNet.
Appendix \ref{app:ensmixmo} analyzes ensembles of Cut-MixMo with CutMix that reach state of the art. 
Finally, we provide a pseudocode in Algorithm \ref{pseudocode}.

\input{sections/sections_appendix/root}%
\input{sections/sections_appendix/features}%
\input{sections/sections_appendix/mheads}%
\input{sections/sections_appendix/expe_details}%
\input{sections/sections_appendix/dynamics}%
\input{pictures/model/different_binarymasks.tex}%
\input{sections/sections_appendix/msda}%
\input{sections/sections_appendix/alpha}%
\input{pictures/experiments/fig_tradeoff_alpha.tex}%
\input{sections/sections_appendix/imagenet.tex}%
\input{tables/table_summary}%
\input{sections/sections_appendix/ensemble}%
\subsection{Pseudo Code}%
Finally, the pseudocode in Algorithm \ref{pseudocode} describes the procedure behind Cut-MixMo with $M=2$.%
\input{pictures/model/pseudocode}%

%% file: sections/sections_appendix/root.tex
\subsection{Weighting function $w_r$}
\label{app:weighting}
As outlined in Section \ref{section:weight}, the asymmetry in the mixing mechanism leads to
asymmetry in the relative importance of the two inputs.
Thus we reweight the loss components with function $w_r$, defined as
$w_{r}(\kappa)=2\frac{\kappa^{1/r}}{\kappa^{1/r} + (1-\kappa)^{1/r}}$. It
rescales the mixing ratio $\kappa$  through the use of a $\frac{1}{r}$ root
operator. In the main paper, we have focused on $r=3$.

Fig.~\ref{fig:wr_plots} illustrates how $w_r$ behaves for $r\in\{1,2,3,4,10\}$
and $r\to\infty$. The first extreme $r=1$ matches the diagonal
$w_{r}(\kappa)=2\kappa$, without rescaling of $\kappa$, similarly to what is customary in MSDA.
Our experiments in Section \ref{expe:hyperparamablation} justified the initial idea to shift the weighting function closer to the
horizontal and constant curve $w_{r}(\kappa)=1$ with higher $r$.
In the other experiments, we always set $r=3$.
\input{pictures/experiments/plot_roots.tex}

%% file: pictures/experiments/plot_roots.tex
\begin{figure}[!h]%
\centering%
\includegraphics[width=\columnwidth]{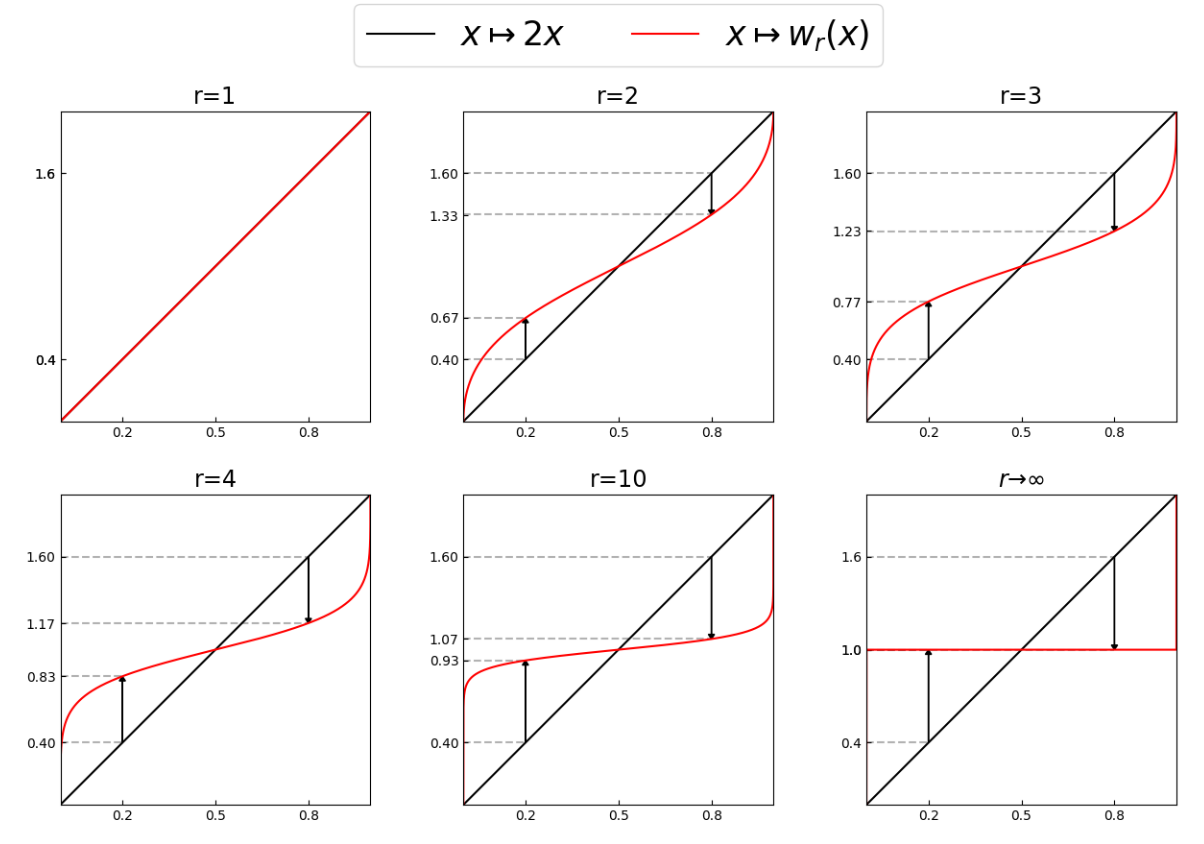}%
\caption{\textbf{Curves of the reweighting operation} that projects $\kappa$ to the flattened ratio $w_r(\kappa)$}%
\label{fig:wr_plots}%
\end{figure}%

%% file: sections/sections_appendix/features.tex
\subsection{Filters activity}%
\label{app:feature}
We argued in Section \ref{sec:msdamixmo} that MixMo better leverages additional parameters in wider networks.
Concretely, a larger proportion of filters in large
networks really help for classification as demonstrated in
Fig.~\ref{fig:mixmo_proj} and \ref{fig:active_feats} in the main paper.
Following common practices in the structured pruning literature
\cite{li2017pruning}, we used the $l_1$-norm of convolutional filters as a proxy
for importance. These 3D filters are of shape $n_{i}\times k \times k$ with
$n_i$ the number of input channels and $k$ the kernel size. In
Fig.~\ref{fig:active_feats}, we arbitrarily defined a filter as active if its
$l_1$-norm is at least $40\%$ of the highest filter $l_1$-norm in that
filter's layer. We report the average percentage of active filters across all filters in the core network $\mathcal{C}$, for 3 learning strategies: vanilla, CutMix and Cut-MixMo.%

The threshold $t_a=0.4$ was chosen for visualization purposes. Nevertheless, the
observed trend in activity proportions remains for varying thresholds in
Tab.~\ref{table:feat_l1}. For example, for the lax $t_a=0.2$, CutMix uses $93.5\%$ of filters vs.\ $98.5\%$ for Cut-MixMo.%
\input{tables/ablation_feat_var}%

%% file: tables/ablation_feat_var.tex
\begin{table}[!t]
  \caption{\textbf{Proportion (\%) of active filters} in core network vs.\ width $w$ for a WRN-28-$w$ on CIFAR 100 and different activity thresholds $t_a$.}
\centering
\resizebox{0.8\linewidth}{!}{%
        \begin{tabular}{c c | c c c c}
            \toprule
          Method & Width & $t_a=0.2$ & $t_a=0.3$ & $t_a=0.4$ & $t_a=0.5$ \\
          \midrule
          \multirow{7}{*}{Vanilla} & 2 & 98.9 & 98.8 & 97.8 & 93.3 \\
           & 3 & 97.3 & 96.4 & 93.2 & 87.5 \\
           & 4 & 96.5 & 95.2 & 91.2 & 81.6 \\
           & 5 & 95.1 & 91.7 & 85.7 & 73.3 \\
           & 7 & 92.6 & 88.2 & 81.0 & 69.5 \\
           & 10 & 87.8 & 80.4 & 71.5 & 57.3 \\
                 & 14 & 83.9 & 74.0 & 61.6 & 46.8 \\ \midrule
          \multirow{7}{*}{CutMix} & 2 & 99.2 & 99.0 & 97.8 & 95.3 \\
                 & 3 & 98.7 & 98.5 & 97.2 & 93.4 \\
                 & 4 & 98.1 & 97.4 & 94.0 & 87.3 \\
                 & 5 & 97.0 & 96.1 & 90.7 & 80.6 \\
                 & 7 & 95.8 & 94.0 & 86.2 & 74.6 \\
                 & 10 & 93.5 & 88.4 & 81.3 & 67.0 \\
                 & 14 & 89.4 & 81.9 & 70.3 & 50.9 \\ \midrule
          \multirow{7}{*}{Cut-MixMo} & 2 & 100.0 & 100.0 & 99.4 & 97.3 \\
           & 3 & 99.8 & 99.8 & 99.7 & 98.7 \\
           & 4 & 99.7 & 99.7 & 99.6 & 98.7 \\
           & 5 & 99.3 & 99.3 & 98.9 & 97.4 \\
           & 7 & 98.9 & 98.8 & 98.0 & 95.2 \\
           & 10 & 98.5 & 98.2 & 96.8 & 92.4 \\
           & 14 & 97.5 & 96.3 & 93.1 & 82.6 \\  \bottomrule
        \end{tabular}
    }
\label{table:feat_l1}
\end{table}

%% file: sections/sections_appendix/mheads.tex
\subsection{Generalization to $M>2$ heads}
\label{app:mheads}

We have mostly discussed our MixMo framework with $M=2$ subnetworks. For better
readability, we referred to the mixing ratios $\kappa$
and $1-\kappa$ with $\kappa \sim \text{Beta}(\alpha, \alpha)$. It's equivalent to a more
generic formulation $(\kappa_0, \kappa_1)\in
\text{Dir}_{2}(\alpha)$ from a symmetric Dirichlet distribution with concentration parameter $\alpha$.
This leads to the alternate equations
$\mathcal{L}_{\text{MixMo}}=\sum_{i=0,1}w_{r}(\kappa_i)\mathcal{L}_{\scalemath{0.59}{\text{CE}}}\left(y_{i},\hat{y}_{i}\right)$,\linebreak
where $w_{r}(\kappa_i)=2\frac{\kappa_i^{1/r}}{\sum_{j=0,1}\kappa_j^{1/r}}.$%

Now generalization to the general case $M\geq 2$ is straightforward. We draw a
tuple $\{\kappa_i\}_{0 \leq i < M}\sim \mbox{Dir}_{M}(\alpha)$ and optimize the
training loss: 
\begin{equation}
\mathcal{L}_{\text{MixMo}}=\sum_{i=0}^{M-1}w_{r}(\kappa_i)\mathcal{L}_{\scalemath{0.59}{\text{CE}}}\left(y_{i},\hat{y}_{i}\right),
\end{equation}
where the new weighting naturally follows:
\begin{equation}%
w_{r}(\kappa_i) = M\frac{\kappa_i^{1/r}}{\sum_{j=0}^{M-1} \kappa_j^{1/r}}, \forall i\in\{0,\dots, M-1\}.
\end{equation}%

The remaining point is the generalization of the mixing block $\mathcal{M}$,
that relies on the existence of MSDA methods for $M>2$ inputs. The linear
interpolation can be easily expanded as in Mixup:
\begin{equation}%
\scalemath{0.9}{\mathcal{M}_{\text{Linear-MixMo}}}\left(\{l_{i}\}\right) = M\sum_{i=0}^{M-1}\kappa_i l_{i},
\end{equation}%
where $l_{i}=c_{i}(x_i)$. However, extensions for other masking MSDAs have only
recently started to emerge \cite{kim2021comixup}. For example, CutMix is not
trivially generalizable to $M>2$, as the patches could overlap and hide
important semantic components. In our experiments, a soft extension of Cut-MixMo performs best: it first linearly interpolates $M-1$ inputs and then patches a region from the $M$-th:
\begin{equation}
\begin{split}
\mathcal{M}_{\text{Cut-MixMo}}\left(\{l_{i}\}\right) = M[
\mathbb{1}_{\mathcal{M}}&\scalemath{1.0}{\odot}l_{k} + \\
(\mathbb{1}-\mathbb{1}_{\mathcal{M}})&\scalemath{1.0}{\odot}\sum_{i=0, i\ne k}^{M-1}\frac{\kappa_i}{1-\kappa_{k}}l_{i}],
\end{split}
\end{equation}
where $\mathbb{1}_{\mathcal{M}}$ is a rectangle of area ratio $\kappa_{k}$ and $k$ sampled
uniformly in $\{0, 1, \dots, M-1\}$. However, it has been less successful than
$M=2$, as only two subnetworks can fit independently in standard parameterization regimes. Future work could design new framework components, such as specific mixing blocks, to tackle these limits.%

%% file: sections/sections_appendix/expe_details.tex
\subsection{Implementation details}
\label{app:expe}

We first used the popular image classification \textbf{datasets} CIFAR-100 and
CIFAR-10 \cite{krizhevsky2009learning}. They contain 60k $32\times32$ natural
and colored images in respectively $100$ classes and $10$ classes, with 50k
training images and 10k test images. At a larger scale, we study Tiny ImageNet
\cite{chrabaszcz2017}, a downsampled version of ImageNet \cite{imagenet_cvpr09}.
It contains $200$ different categories, 100k $64\times64$ training images
(\textit{i.e.}\ 500 images per class) and 10k test images.%

Our code was adapted from the official MIMO \cite{havasi2020raining}
implementation\footnote{\url{https://github.com/google/edward2/}}. For CIFAR,
we re-use the \textbf{hyper-parameters} from MIMO \cite{havasi2020raining}. The
optimizer is SGD with learning rate of $\frac{0.1}{b}\times\frac{
\text{batch-size}}{128}$, batch size $64$, linear warmup over 1 epoch, decay
rate 0.1 at steps $\{100, 200, 225\}$, $l_2$ regularization 3e-4. We follow
standard MSDA practices \cite{ashukha2020pitfalls,kim2020puzzle,yun2019cutmix} and
set the maximum number of epochs to $300$. For Tiny ImageNet, we adapt
PreActResNet-18-$w$, with $w\in\{1,2,3\}$ times more filters. We re-use the
hyper-parameters from Puzzle-Mix \cite{kim2020puzzle}. The optimizer is SGD with
learning rate of $\frac{0.2}{b}$, batch size $100$, decay rate $0.1$ at steps
$\{600, 900\}$, $1200$ epochs maximum, weight decay 1e-4.
Our experiments ran on a single NVIDIA 12Go-TITAN X Pascal GPU.
All results without a $\dagger$ were obtained with these training configurations.
We will soon release our code and pre-trained models to facilitate reproducibility.

\textbf{Batch repetition} increases performances at the cost of \textbf{longer training}, which may be discouraging for some practitioners. Thus in addition to $b=4$ as in MIMO \cite{havasi2020raining}, we often consider the quicker $b=2$. Note that most of our concurrent approaches also increase training time: DE
\cite{lakshminarayanan2016simple} via several independent trainings, Puzzle-Mix
\cite{kim2020puzzle} via saliency detection ($\approx\times2$), GradAug
\cite{yang2020radaug} via multiple subnetworks predictions ($\approx\times3$) or
Mixup BA \cite{Hoffer_2020_CVPR} via $10$ batch augmentations ($\approx\times7$
with our hardware on a single GPU).%

MixMo operates in the features space and is complementary with \textbf{pixels
augmentations}, \textit{i.e.}\ cropping, AugMix. The standard vanilla pixels
data augmentation \cite{he51deep} consists of $4$ pixels padding, random
cropping and horizontal flipping. When combined with CutMix, notably to benefit
from multilabel smoothing, the input may be of the form: $\left(m_x(x_{i}, x_{k},
\lambda), x_{j}\right)$, where $x_k$ is randomly chosen in the whole dataset, and not
only inside the current batch\footnote{Following \url{https://github.com/ildoonet/cutmix}}. Moreover,
$\mathcal{M}_{\text{Cut-MixMo}}$ modifies by $\mathbb{1}_{\mathcal{M}}$ the
visible part from mask $\mathbb{1}_m$ (of area $\lambda$). We thus modify
targets accordingly: $(\lambda' y_{i} + (1-\lambda')y_{k}, y_{j})$ where
$\lambda' = \frac{\sum \mathbb{1}_m \odot \mathbb{1}_{\mathcal{M}}}{\sum
\mathbb{1}_{\mathcal{M}}}$. To fully benefit from $b$, we force the repeated
$x_{i}$ to remain predominant in its $b$ appearances: \textit{i.e.}, we swap
$x_{i}$ and $x_{k}$ if $\lambda'<0.5$. We see CutMix as a perturbation on the
main batch sample.

Distributional uncertainty measures help when there is a mismatch between train
and test data distributions. Thus \cite{hendrycks2018benchmarking} introduced
\textbf{CIFAR-100-c} on which \textbf{AugMix} performs best. AugMix sums the
pixels from a chain of several augmentations and is complementary to our
approach in features. We use default
parameters\footnote{\url{https://github.com/google-research/augmix/blob/master/cifar.py}}:
the severity is set 3, the mixture's width to 3 and the mixture's depth to 4. We
exclude operations in AugMix which overlap with CIFAR-100-c corruptions: thus,
[equalize, posterize, rotate, solarize, shear\_x, shear\_y, translate\_x,
translate\_y] remain. We disabled the Jensen-Shannon Divergence loss between
predictions for the clean image and for the same image AugMix
augmented: that would otherwise triple the training time. For comparison of out-of-domain uncertainty estimations, we report NLL as in \cite{havasi2020raining,ovadia2019can}: indeed, the recommendation of \cite{ashukha2020pitfalls} to apply TS only stands for in-domain test set.

\input{tables/main_cifar10100_end.tex}
\input{pictures/experiments/training_dynamics_p.tex}%

\subsection{Evaluation setting and metrics}
\label{app:expeeval}
We reproduce the \textbf{experimental setting} from CutMix
\cite{yun2019cutmix}, Manifold Mixup \cite{manifoldmixup19} and other works such
as the recent state-of-the-art ResizeMix \cite{qin2020resizemix}: in absence of
a validation dataset, results are reported at the epoch that yields the best
test accuracy. For fair comparison, we apply this early stopping for all concurrent
approaches. Nonetheless, for the sake of completeness, Table \ref{table:cifar10010end}
shows results without early stopping on the main experiment (CIFAR with a
standard WRN-28-10). We recover the exact same ranking among methods as in Table \ref{table:cifar10010}.

Following recent works in ensembling \cite{chirkova2020deep,lobacheva2020power,rame2021dice}, we have mainly focused on the NLL$_c$ \textbf{metric} for in-domain test set. Indeed, \cite{ashukha2020pitfalls} have shown that \enquote{comparison of \textelp{} ensembling methods without temperature scaling (TS) \cite{guo2017calibration} might not provide a fair ranking}. Nevertheless  in Table \ref{table:cifar10010end}, we found that \textit{Negative Log-Likelihood} (NLL) (without TS) leads to similar conclusions as NLL$_c$ (after TS).

The TS even mostly seems to benefit to poorly calibrated models, as shown by the calibration criteria \textit{Expected Calibration Error} (ECE, $\downarrow$, 15 bins). ECE measures how confidences match accuracies. MixMo attenuates over-confidence in large networks and thus reduces ECE. In our case, combining ensembling and data augmentation improves calibration \cite{wen2021combining}.
Note that the appropriate measure of calibration is still under debate \cite{nixon2019measuring}.
Notably, \cite{ashukha2020pitfalls} have also stated that, despite being widely used, ECE is biased and unreliable: we can confirm that we found ECE to be dependant to hyper-parameters and implementation details. Due to space constraints and these pitfalls, we have not included this controversial metric in the main paper.

%% file: tables/main_cifar10100_end.tex
\begin{table}[!t]
    \caption{\textbf{WRN-28-10 on CIFAR} without early stopping.}%
    \centering
    \resizebox{1.0\linewidth}{!}{%
    \begin{tabular}{c | c | c c c c c | c c c c}
        \toprule
        \multicolumn{2}{c|}{Dataset} & \multicolumn{5}{c|}{CIFAR-100} & \multicolumn{4}{c}{CIFAR-10} \\
        \midrule
        \midrule
        Approach & \begin{tabular}{@{}c@{}}\scalebox{1.0}{Time}\\\scalebox{1.0}{Tr./Inf.}\end{tabular} & \begin{tabular}{@{}c@{}}\scalebox{1.0}{Top1}\\\scalebox{0.9}{$\%, \uparrow$}\end{tabular} & \begin{tabular}{@{}c@{}}\scalebox{1.0}{Top5}\\\scalebox{0.9}{$\%, \uparrow$}\end{tabular} & \begin{tabular}{@{}c@{}}\scalebox{1.0}{NLL$_{c}$}\\\scalebox{0.9}{$10^{-2}, \downarrow$}\end{tabular} & \begin{tabular}{@{}c@{}}\scalebox{1.0}{NLL}\\\scalebox{0.9}{$10^{-2}, \downarrow$}\end{tabular} & \begin{tabular}{@{}c@{}}\scalebox{1.0}{ECE}\\\scalebox{0.9}{$10^{-2}, \downarrow$}\end{tabular} & \begin{tabular}{@{}c@{}}\scalebox{1.0}{Top1}\\\scalebox{0.9}{$\%, \uparrow$}\end{tabular} & \begin{tabular}{@{}c@{}}\scalebox{1.0}{NLL$_{c}$}\\\scalebox{0.9}{$10^{-2}, \downarrow$}\end{tabular} & \begin{tabular}{@{}c@{}}\scalebox{1.0}{NLL}\\\scalebox{0.9}{$10^{-2}, \downarrow$}\end{tabular} & \begin{tabular}{@{}c@{}}\scalebox{1.0}{ECE}\\\scalebox{0.9}{$10^{-2}, \downarrow$}\end{tabular}\\
        \midrule
        \midrule
        Vanilla & \multirow{4}{*}{1/1} & 81.47 & 95.57 & 73.6 & 76.2 & 6.47 & 96.31 & 12.5 & 14.1 & 1.95 \\
        Mixup & & 83.15 & 95.75 & 66.3 & 67.3 & \textbf{1.62} & 97.00 & 11.3 & 11.5 & 0.97 \\
        Hard PatchUp$^\dagger$ & & 83.87 & - & - & 66.0 & - & 97.47 & - & 11.4 & - \\
        CutMix & & 83.74 & 96.18 & 65.4 & 66.1 & 4.95 & 97.21 & 9.7& 10.8 & 1.51 \\
        \midrule
        Puzzle-Mix$^\dagger$ & 2/1 & 84.05 & 96.08& 66.9 & 68.1 & 2.76 & - & -& - & - \\
        \midrule
        GradAug$^\dagger$ & \multirow{2}{*}{3/1} & 83.98 & 96.28 & -& - & - & - & -& - & - \\
        + CutMix$^\dagger$ & & 85.25 & 96.85 & -& - & - & - & -& - & - \\
        \midrule
        Mixup BA$^\dagger$ & 7/1 & 84.30 & - & - & - & - &\textbf{97.80} &- & - &- \\
        \midrule
        DE (2 Nets) & \multirow{2}{*}{2/2} & 83.15 & 96.30 & 66.0 & 67.2 & 5.15 & 96.58 & 11.1& 12.2 & 1.82 \\
        + CutMix & & 85.46 & 96.90 & 57.4& 57.5 & 3.62 & 97.51 & 8.7& 9.0 & 1.16 \\
        \midrule
        \midrule
        MIMO ($M=2$) & \multirow{5}{*}{2/1} & 82.04 & 95.75 & 69.1& 72.4 & 6.32 & 96.33 & 12.1& 13.4 & 1.89 \\
        \cmidrule{1-1}
        \cmidrule{3-11}
        \multirow{1}{*}{Linear-MixMo} & & 81.88 & 95.97 & 67.8& 70.3 & 6.20 & 96.55 & 11.4 & 12.5 & 1.67 \\
        + CutMix & & 84.55 & 96.95 & 57.4& 57.5 & 2.54 & 97.34 & 8.9& 9.3 & 1.34 \\
        \cmidrule{1-1}
        \cmidrule{3-11}
        \multirow{1}{*}{Cut-MixMo} & & 84.07 & 96.97 & 56.6& 57.9 & 4.19 & 97.26 & 8.7& 9.1 & 0.98 \\
        + CutMix & & 85.17 & 97.28 & 54.4& 54.5 & 2.13 & 97.33 & 8.5& 8.6 & \textbf{0.88} \\
        \midrule
        MIMO ($M=2$) & \multirow{6}{*}{4/1} & 82.74 & 95.90 & 67.0& 74.0 & 7.56 & 96.66 & 11.5& 13.6 & 1.98 \\
        MIMO$^\dagger$ ($M=3$)& & 82.0 & - & -& 69.0 & 2.2 & 96.4 & -& 12.3 & 1.0 \\
        \cmidrule{1-1}
        \cmidrule{3-11}
        \multirow{1}{*}{Linear-MixMo} & & 82.53 & 96.08 & 65.8& 68.5 & 6.64 & 96.78 & 10.8& 11.8 & 1.80 \\
        + CutMix & & 85.24 & 96.97 & 56.3& 56.4 & 3.53 & 97.53 & 8.8& 8.6 & 1.19 \\
        \cmidrule{1-1}
        \cmidrule{3-11}
        \multirow{1}{*}{Cut-MixMo} &  & 85.32 & 97.12 & 53.6& 54.8 & 4.53 & 97.42 & 8.1& 8.4 & 1.15 \\
        + CutMix & & \textbf{85.59} & \textbf{97.33} & \textbf{53.2} & \textbf{53.3} & 1.95 & 97.70 & \textbf{8.0} & \textbf{8.2} & 0.98 \\
        \bottomrule%
    \end{tabular}%
    }%
    \label{table:cifar10010end}%
    \end{table}%

%% file: pictures/experiments/training_dynamics_p.tex
\begin{figure*}[!ht]
\centering
\includegraphics[width=0.78\linewidth]{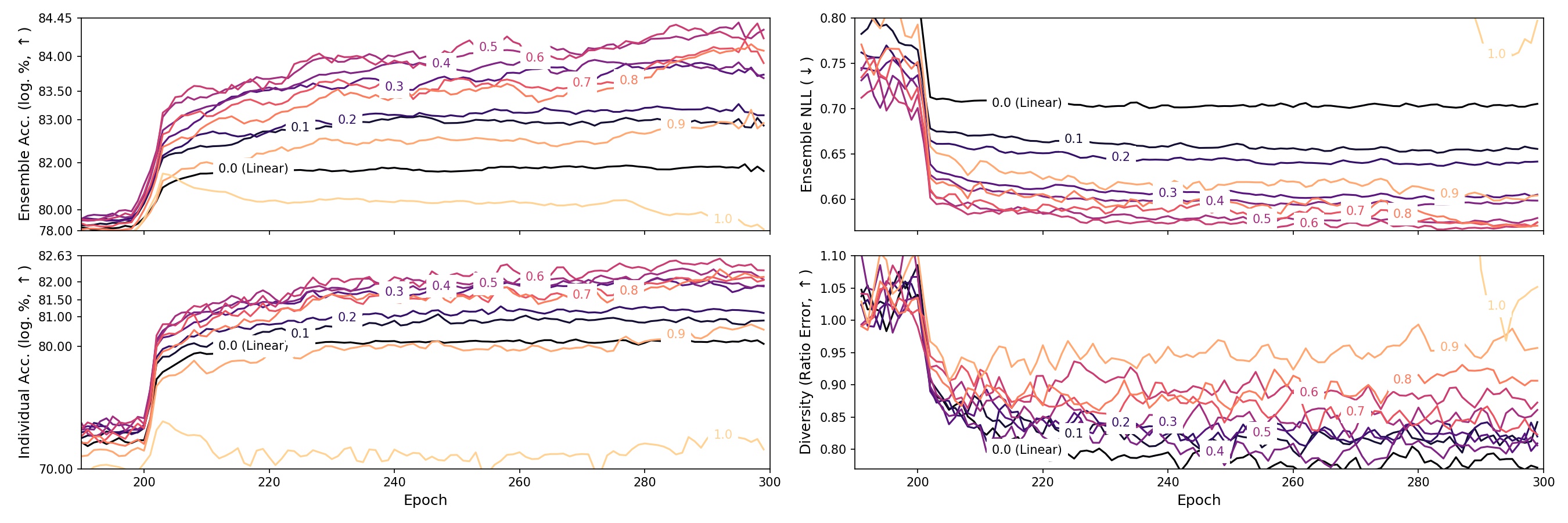}
\caption{\textbf{Training dynamics}. Higher probability $p$ of binary mixing via patches increases diversity (lower right), and also subnetworks accuracy (lower left) but only up to $p=0.6$. Around this value, we obtain best ensemble performances, in terms of accuracy (upper left) or uncertainty estimation (upper right). $b=2$, $r=3$, $\alpha=3$ with WRN-28-10 on CIFAR-100.}
\label{fig:dynamicsp}
\end{figure*}

%% file: sections/sections_appendix/dynamics.tex
\subsection{Training dynamics}
\label{app:td}

Fig.~\ref{fig:dynamicsp} showcases training dynamics for probability $p\in[0,1]$
of patch mixing (see Section \ref{expe:mixingmechanism}). In the remaining $1-p$, we interpolate features linearly. For $p=0$, we recover our Linear-MixMo; for $p=0.5$, we recover our Cut-MixMo. In all approaches, $p$ is linearly reduced towards $0$ beyond the $\frac{11}{12}$ of the training epochs, \textit{i.e.}\ from epoch 275 to 300 on CIFAR. As we sum at inference, this reduces the train-test distribution gap and slightly increases individual accuracy during the final epochs (lower left in Fig.~\ref{fig:dynamicsp}).

Diversity is measured by the ratio-error, the ratio between the number of samples on which only one of the two predictor is wrong, divided by the number of samples on which they are both wrong. It is positively correlated with $p$. However, individual accuracies first increase
with $p$ until $p=0.6$, then the tendency is reversed.
Overall, best ensemble performances in terms of accuracy (Top1) and uncertainty (NLL)
estimation are obtained with $p\in[0.5, 0.6]$. Most importantly, we note that the performance gaps are consistent and stable along training.

%% file: pictures/model/different_binarymasks.tex
\begin{figure}[!b]%
\includegraphics[width=\linewidth]{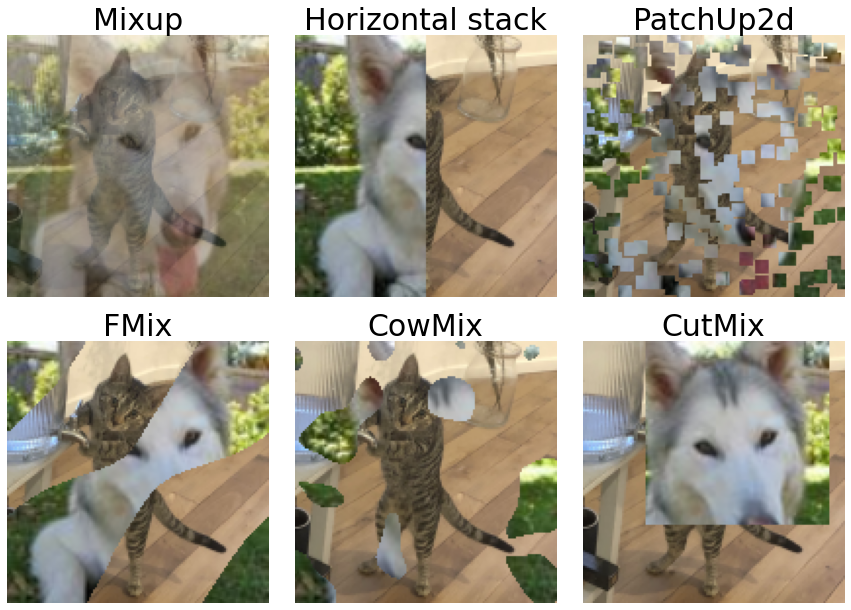}%
\caption{\textbf{Common MSDA procedures with $\lambda=0.5$}.}%
\label{fig:msda}%
\end{figure}%

%% file: sections/sections_appendix/msda.tex
\subsection{Mixed sample data augmentations}
\label{app:msda}
We have drawn inspiration from MSDA techniques to design our mixing block
$\mathcal{M}$. In particular, Section \ref{expe:mixingmechanism} compared
different $\mathcal{M}$ based on recent papers. Fig.~\ref{fig:msda}
provides the reader a visual understanding of their behaviour, which we explain
below.%

\textbf{MixUp} \cite{zhang2018mixup} linearly interpolates between pixels
values: $m_x(x_i,x_k,\lambda) = \lambda x_i + (1-\lambda) x_k$. The remaining methods
fall under the label of \textbf{binary MSDA}: $m_x(x_{i}, x_{k}, \lambda) =
\mathbb{1}_m \odot x_{i} + (\mathbb{1}-\mathbb{1}_m) \odot x_{k}$ with
$\mathbb{1}_m$ a mask with binary values $\{0,1\}$ and area of ratio $\lambda$.
They diverge in how this mask is created. The \textbf{horizontal
  concatenation}, also found in \cite{summers2019improved}, simply draws a
vertical line such that every pixel to the left belongs to one sample and every
pixel to the right belongs to the other. Similarly, we define a \textbf{vertical
  concatenation} with an horizontal line. \textbf{PatchUp}
\cite{faramarzi2020patchup} adapted DropBlock \cite{ghiasi2018dropblock}: a
canvas $C$ of patches is created by sampling for every spatial coordinate from
the Bernoulli distribution $\text{Ber}(\lambda')$ (where $\lambda'$ is a recalibrated
value of $\lambda$): if the drawn binary value is $1$, a patch around that
coordinate is set to $1$ on the final binary mask $\mathbb{1}_m$. PatchUp was
designed for in-manifold mixing with a different mask by channels. However,
duplicating the same 2D mask in all channels for $\mathcal{M}$ performs better in our
experiments. \textbf{FMix} \cite{harris2020mix} selects a large contiguous
region in one image and pastes it onto another. The binary mask is made of the
top-$\lambda$ percentile of pixels from a low-pass filtered 2D map $G$ drawn
from an isotropic Gaussian distribution. \textbf{CowMix}
\cite{french2019semi,french2020milking} selects a cow-spotted set of regions,
and is somehow similar to FMix with a Gaussian filtered 2D map $G$.
\textbf{CutMix} \cite{yun2019cutmix} was inspired by CutOut
\cite{devries2017improved}. Formally, we sample a square with edges of length
$R\sqrt{\lambda}$, where $R$ is the length of an image edge. Note that this
sometimes leads to non square rectangles when the initially sampled square
overlaps with the edge from the original image. We adjust our $\lambda$ a
posteriori to fix this boundary effect. Regarding the hyper-parameters, we
use in $\mathcal{M}$ those provided in the seminal papers, except for sampling of $\kappa$ where we set $\alpha=2$ in all setups.%

Note we consider both versions of MixUp (in-pixel and manifold) in this paper, but only the in-pixel version of CutMix. Indeed, the manifold version of CutMix was shown in the seminal CutMix paper \cite{yun2019cutmix} to be inferior to the standard in-pixel variant.

%% file: sections/sections_appendix/alpha.tex
\subsection{Hyper-parameter $\alpha$}
\label{app:alpha}
In Fig.~\ref{fig:tradeoff_alpha}, we study the impact of different values of
$\alpha$, parameterizing the sampling law for $\kappa \sim \text{Beta}(\alpha,\alpha)$.
For high values of $\alpha$, the interval of $\kappa$ narrows down around $0.5$.
Diversity is therefore decreased: we speculate this is because we do not benefit
anymore from lopsided updates. The opposite extreme, when $\alpha\scalemath{1.}{=}1$, is
equivalent to uniform distribution between $0$ and $1$. Therefore diversity is
increased, at the cost of lower individual accuracy due to less stable training.
For simplicity, we set $\alpha\scalemath{1.}{=}2$. Manifold-Mixup \cite{manifoldmixup19} selected the same value on CIFAR-100. However, this value could be fine tuned on the target task: \textit{e.g.}\ in
Fig.~\ref{fig:tradeoff_alpha}, $\alpha\scalemath{1.}{=}4$ seems to perform best for Cut-MixMo on CIFAR-100 with WRN-28-10 with $r\scalemath{1.}{=}3$, $p\scalemath{1.}{=}0.5$ and $b\scalemath{1.}{=}2$.%

%% file: pictures/experiments/fig_tradeoff_alpha.tex
\begin{figure}[!h]
\centering%
\includegraphics[width=0.855\linewidth]{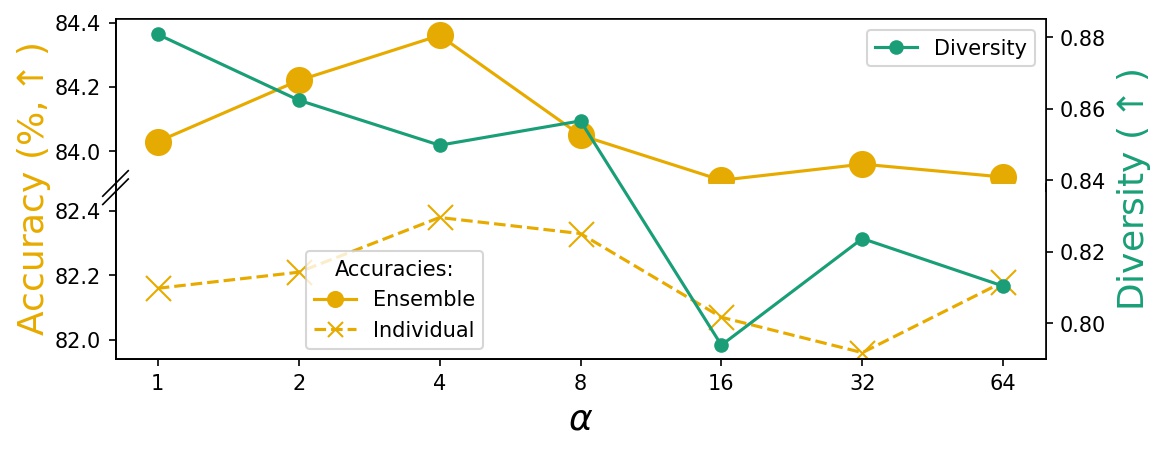}%
\vspace{-0.5em}%
\caption{\textbf{Diversity/accuracy} as function of $\alpha$.}%
\label{fig:tradeoff_alpha}%
\end{figure}%

%% file: sections/sections_appendix/imagenet.tex
\subsection{Preliminary ImageNet experiments}
\label{app:imagenet}

To further prove MixMo's ability to scale to more complex problems, we also conduct a preliminary study of its behavior on the larger scale ImageNet dataset \cite{imagenet_cvpr09}. Following the protocol outlined in the seminal MIMO paper \cite{havasi2020raining}, we consider variations on the standard ResNet-18 in the form of ResNet-18-$w$ networks where $w$ is multiplicative width factor.

These first experiments confirm that MixMo performs well when networks are overparameterized. For values of $w\geq 5$, our network at the end of training outperforms both Vanilla and CutMix baselines. For example, with a ResNet-18-5 backbone, Cut-MixMo (78.20\% Top1, 0.867 NLLc) improves over Vanilla (76.47\%, 1.121) and CutMix (77.40\%, 1.263). This remains the case for a ResNet-18-7 backbone with Cut-MixMo (78.55\% Top1, 0.846 NLLc) outperforming Vanilla (76.86\%, 1.100) and CutMix (77.18\%, 1.190).

%% file: tables/table_summary.tex
\begin{table*}[!h]%
\caption{\textbf{Summary}: WRN-28-$w$ on CIFAR-100. $b=4$.}
\centering
\resizebox{0.85\linewidth}{!}{%
\begin{tabular}{c | c | c c | c c| c c | c c | c c }
    \toprule
    Width & \multirow{1}{*}{Approach}  & \multicolumn{2}{c|}{1-Net} &  \multicolumn{2}{c|}{2-Nets} & \multicolumn{2}{c|}{Linear-MixMo} & \multicolumn{2}{c|}{Cut-MixMo} & \multicolumn{2}{c}{2-Cut-MixMos} \\
    $w$ &  CutMix & - & \checkmark & - & \checkmark & - & \checkmark & - & \checkmark & - & \checkmark \\
    \midrule
    \midrule
    \multirow{3}{*}{2} & Top1 & 76.44 & 78.06 & 79.16 & 80.81 & 75.82 & 76.36 & 75.66 & 75.17 & 76.98 & 76.11 \\
    & NLL$_c$ & 0.921 & 0.815 & 0.776 & 0.695 & 0.841 & 0.824 & 0.824 & 0.846 & 0.7661 & 0.798\\
    & \# params & \multicolumn{2}{c|}{1.48M} & \multicolumn{2}{c|}{2.95M} & \multicolumn{4}{c|}{1.49M} & \multicolumn{2}{c}{2.99M} \\
    \midrule
    \multirow{3}{*}{3} & Top1 & 77.95 & 80.70 & 80.85 & 83.14 & 78.51 & 80.74 & 79.81 & 79.85 & 80.78 & 81.20 \\
    & NLL$_c$ & 0.862 & 0.750 & 0.738 & 0.644 & 0.760 & 0.696 & 0.693 & 0.702 & 0.635 & 0.650\\
    & \# params & \multicolumn{2}{c|}{3.31M} & \multicolumn{2}{c|}{6.62M} & \multicolumn{4}{c|}{3.33M} & \multicolumn{2}{c}{6.66M} \\
    \midrule
    \multirow{3}{*}{4} & Top1 & 78.84 & 81.55 & 81.48 & 83.93 & 80.43 & 81.66 & 81.68 & 81.69 & 82.57 & 82.58\\
    & NLL$_c$ & 0.824 & 0.711 & 0.711 & 0.609 & 0.712 & 0.656 & 0.646 & 0.635 & 0.590 & 0.588\\
    & \# params & \multicolumn{2}{c|}{5.87M} & \multicolumn{2}{c|}{11.74M} & \multicolumn{4}{c|}{5.89M} & \multicolumn{2}{c}{11.79M} \\
    \midrule
    \multirow{3}{*}{5} & Top1 & 79.75 & 82.55 & 82.18 & 84.60 & 80.95 & 83.06 & 83.11 & 83.34 & 83.97 & 84.31\\
    & NLL$_c$ & 0.813 & 0.686 & 0.693 & 0.596 & 0.703 & 0.617 & 0.598 & 0.591 & 0.549 & 0.546\\
    & \# params & \multicolumn{2}{c|}{9.16M} &  \multicolumn{2}{c|}{18.32M} & \multicolumn{4}{c|}{9.19M} &  \multicolumn{2}{c}{18.39M} \\
    \midrule
    \multirow{3}{*}{7} & Top1 & 81.14 & 83.71 & 82.94 & 85.52 & 82.4 & 84.51 & 84.32 & 84.94 & 85.50 & 85.90 \\
    & NLL$_c$ & 0.764 & 0.648 & 0.673 & 0.573 & 0.675 & 0.581 & 0.562 & 0.543 & 0.516 & 0.498 \\
    & \# params & \multicolumn{2}{c|}{17.92M} & \multicolumn{2}{c|}{35.85M} & \multicolumn{4}{c|}{17.97M}  &  \multicolumn{2}{c}{35.94M} \\
    \midrule
    \multirow{3}{*}{10} & Top1 & 81.63 & 84.05& 83.17 & 85.74 & 83.08 & 85.47 & 85.40 & 85.77 & 86.04 & 86.63 \\
    & NLL$_c$ & 0.750 & 0.644 & 0.668 & 0.571 & 0.656 & 0.558 & 0.535 & 0.524 & 0.494 & 0.479 \\
    & \# params & \multicolumn{2}{c|}{36.53M} & \multicolumn{2}{c|}{73.07M} & \multicolumn{4}{c|}{36.60M} & \multicolumn{2}{c}{73.21M} \\
    \midrule
    \multirow{3}{*}{14} & Top1 & 82.01 & 84.31 & 83.47 & 85.80 & 83.79 & 86.05 & 85.76 & 86.19 & 86.58 & 87.11  \\
    & NLL$_c$ & 0.730 & 0.645 & 0.656 & 0.569 & 0.648 & 0.545 & 0.527 & 0.518 & 0.488 & 0.473 \\
    & \# params & \multicolumn{2}{c|}{71.55M} & \multicolumn{2}{c|}{143.1M} & \multicolumn{4}{c|}{71.64M} & \multicolumn{2}{c}{143.28M} \\
    \bottomrule
\end{tabular}
    }
\label{table:splitadvantage}
\end{table*} 

%% file: sections/sections_appendix/ensemble.tex
\input{pictures/experiments/fig_smacx}
\subsection{Ensemble of Cut-MixMo with CutMix}
\label{app:ensmixmo}
Fig.~\ref{fig:smacx} 
plots performance for different
widths $w$ in WRN-28-$w$ and varying number of ensembled networks $N$: two
vertically aligned points have the same parameter budget. Indeed, the total number of
parameters in our architectures has been used as a proxy for model complexity, as in \cite{chirkova2020deep,lobacheva2020power}.
The increase in the total number of weights in MixMo is visually almost unnoticeable. Precisely, with WRN-28-10, MixMo ($M$=$2$) has 36.60M weights vs. 36.53M standardly (+0.2\%). Moreover, the number of flops is 5.9571G Flops for MixMo vs. 5.9565G Flops standardly (+0.01\%). That's why we state we achieve
ensembling (almost) ``for free''.

We compare
ensembling with CutMix rather than standard pixels data augmentation, as previously done in Fig.~\ref{fig:batchrepet} from Section \ref{expe:ensemblemixmo}. CutMix induces additional regularization and label smoothing: empirically, it improves all our approaches. For a
fixed memory budget, a single network usually performs worse than an ensemble of
several medium-size networks: we recover the \textbf{Memory Split Advantage} even with CutMix.
However, Cut-MixMo challenges this by remaining closer to the
lower envelope. In other words,
parameters allocation (more networks or bigger networks) has less impact on
results. This is due to Cut-MixMo's ability to better use large networks.

In Table \ref{table:splitadvantage}, we summarize several experiments on
CIFAR-100. Among other things, we can observe that large vanilla networks tend
to gain less from ensembling \cite{lobacheva2020power}: \textit{e.g}.\ 2
vanillas WRN-28-10 ($83.17\%$ Top1, $0.668$ NLL$_c$) do not perform much better
than 2 WRN-28-7 ($82.94\%$, $0.673$). This remains true even with CutMix:
($85.74\%$, $0.571$) vs.\ ($85.52\%$, $0.573$). We speculate this is related to
wide networks' tendency to converge to less diverse solutions, as studied in
\cite{neal2018modern}. Contrarily, \textbf{MixMo improves the ensembling of large networks}, with ($86.04\%$, $0.494$) vs.\ ($85.50\%$, $0.517$) on the same setup. When additionally combined with CutMix, we obtain state of the art
($86.63\%$, $0.479$) vs.\ ($85.90\%$, $0.498$). 
This demonstrates the importance of Cut-MixMo in cooperation with standard pixels data augmentation.
It attenuates the drawbacks from over-parameterization
This is of great importance for practical efficiency: it modifies
the optimal network width for real-world applications.


%% file: pictures/experiments/fig_smacx.tex
\begin{figure}[!b]%
\centering%
\includegraphics[width=1.0\linewidth]{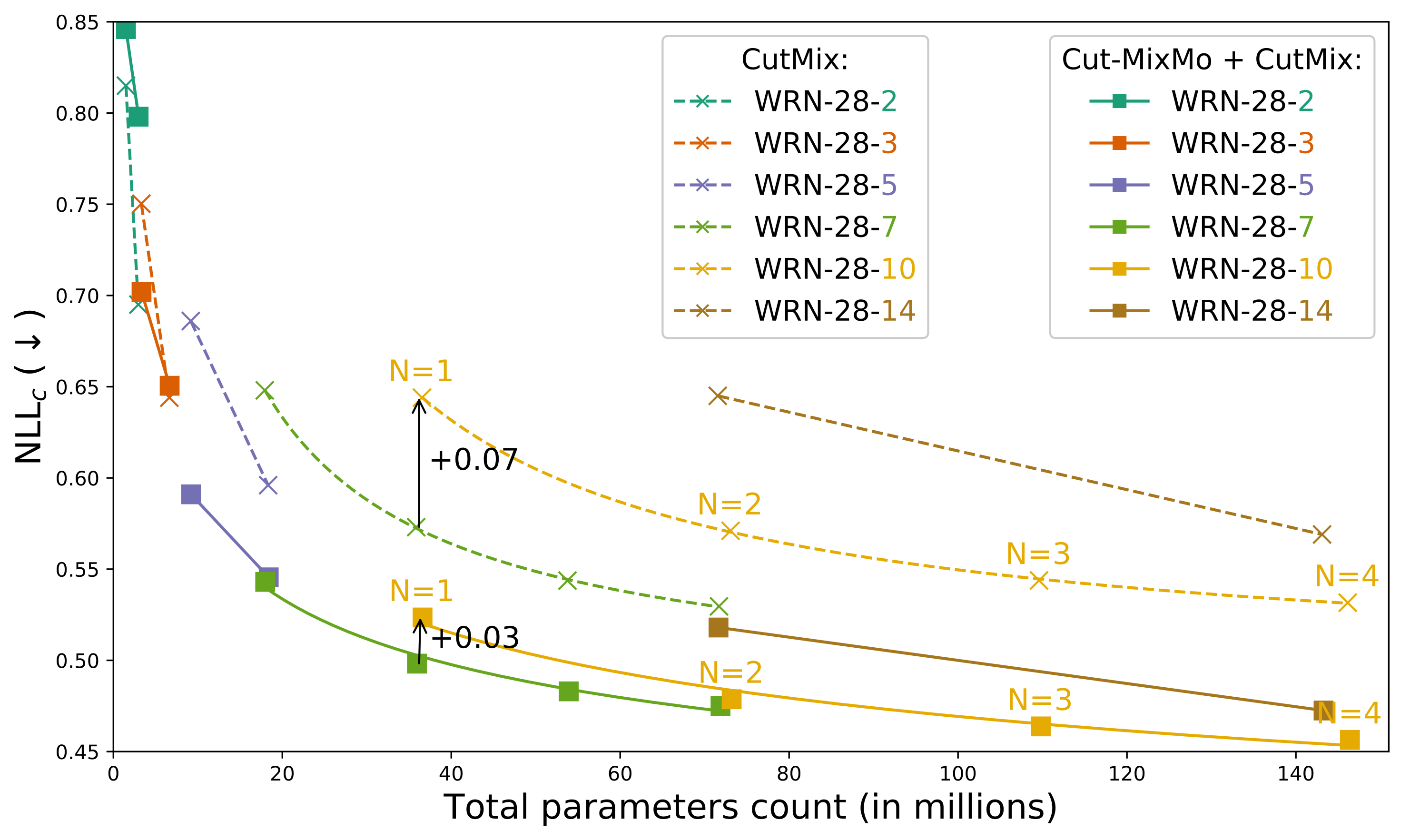}%
\caption{\textbf{Ensemble effectiveness} (NLL$_c$/\#params). We slide the width in WRN-28-$w$ and numbers of members $N$. CutMix data augmentation. Interpolations through power laws \cite{lobacheva2020power} when more than 2 points are available.}%
\label{fig:smacx}%
\end{figure}%

%% file: pictures/model/pseudocode.tex
\begin{algorithm*}
\DontPrintSemicolon
\tcc{Setup}%
\KwParams{First convolutions $\{c_0,c_1\}$, dense layers $\{d_0,d_1\}$ and core network $\mathcal{C}$, randomly initialized.}%
\KwInput{Dataset $D=\{x_{i}, y_{i}\}_{i=1}^{|D|}$, probability $p$ of applying binary mixing via patches, reweighting coefficient $r$, concentration parameter $\alpha$, batch size $b_s$, batch repetition $b$, optimizer $g$ , learning rate $l_r$.}%
\tcc{Training Procedure}%
\For{epoch \textbf{from} $1$ \textbf{to} \#epochs} {%
\For{step \textbf{from} $1$ \textbf{to} $\frac{|D|\times b}{b_s}$} {%
\tcc{Step 1: Batch creation}%
\text{Randomly select} $\frac{b_s}{b}$ samples \tcp*{Sampling}%
\text{Duplicate these samples $b$ times} to create batch $\{x_i, y_{i}\}_{i \in B}$ of size $b_s$ \tcp*{Batch repetition}%
\text{Randomly shuffle $B$ with $\pi$} to create $\{(x_i, x_{j}), (y_{i}, y_{j})\}_{i \in B, j=\pi(i)}$ \tcp*{Shuffling}%
\tcc{Step 2: Define the mixing mechanism at the batch level}%
\uIf{$\text{epoch} >\frac{11}{12}\times \text{\#epochs}$}{
$p_e=p\frac{\text{\#epochs} - \text{epoch}}{\frac{1}{12}\times \text{\#epochs}}$ \tcp*{Linear descent to $0$ over the last twelfth of training}
}
\Else{
$p_e=p$
}
Sample $1_{binary} \sim \text{Ber}(p_e)$ from Bernoulli distribution \tcp*{Whether we apply binary or linear mixing}
Sample $1_{outside} \sim \text{Ber}(0.5)$ \tcp*{Whether the first input is inside or outside the rectangle}
\tcc{Step 3: Forward and loss}%
\For{$i\in B$} {%
    Sample $\kappa_i \sim \text{Beta}(\alpha,\alpha)$\\
    $l_i^0 = c_0(x_i)$ and $l_i^1 = c_1(x_{\pi(i)})$\\
  \uIf{$1_{binary}$}{
    Sample $\mathbb{1}_{\mathcal{M}}$ a rectangular binary mask with average $\kappa_i$ (as in CutMix)\\
    \uIf{$1_{outside}$}{
    $\mathbb{1}_{\mathcal{M}} \gets \mathbb{1} - \mathbb{1}_{\mathcal{M}}$ \tcp*{Permute the rectangle and its complementary}
    $\kappa_i \gets 1-\kappa_i$
    }
    $l_i = 2\left[\mathbb{1}_{\mathcal{M}}\scalemath{1.0}{\odot}l_{0}+(\mathbb{1}-\mathbb{1}_{\mathcal{M}})\scalemath{1.0}{\odot} l_{1}\right]$ \tcp*{Apply binary mixing}
  }
   \Else{
    $l_i = 2\left[\kappa_i l_{0} + (1-\kappa_i) l_{1}\right]$ \tcp*{Apply linear interpolation}
  }
Extract features $f_i \gets \mathcal{C}(l_i)$ from core network \\
Compute predictions $\hat{y}_i^0\gets d_0(f_i)$ and $\hat{y}_i^1\gets d_1(f_i)$ \\
Compute weights $w_i \gets 2\frac{\kappa_i^{1/r}}{\kappa_i^{1/r} + (1-\kappa_i)^{1/r}}$\\
Compute loss $\mathcal{L}_i \gets w_{i}\mathcal{L}_{\scalemath{0.59}{\text{CE}}}\left(y_{i},\hat{y}_{i}^0\right)+(2-w_{i})\mathcal{L}_{\scalemath{0.59}{\text{CE}}}\left(y_{\pi(i)}, \hat{y}_{i}^1\right)$
}
Average loss $\mathcal{L}_{\text{MixMo}} \gets \frac{1}{|B|}\sum \mathcal{L}_i $\\
\tcc{Step 4: Back propagation}%
$c_0,c_1,\mathcal{C},d_0,d_1 \gets g\left(\text{gradient}=\nabla \mathcal{L}_{\text{MixMo}}, \text{learning rate}=\frac{l_r}{b}\right)$\\
}}%
\tcc{Test Procedure}%
\KwData{Inputs $\{x_i\}_{i=1}^T$ \tcp*{Test Data}}%
\For{$i \in \{1,\dots, T \}$} {%
\text{Extract features} $f_i = \mathcal{C}\left(c_0(x_i) + c_1(x_i)\right)$ \\%
\KwOutput{$\frac{1}{2} \left[d_0(f_i) + d_1(f_i)\right])$}}%
\caption{Procedure for Cut-MixMo with $M=2$ subnetworks}%
\label{pseudocode}
\end{algorithm*}%